\documentclass{article}

\usepackage{microtype}
\usepackage{graphicx}
\usepackage{subfigure}
\usepackage{booktabs} %
\usepackage[dvipsnames]{xcolor}

\usepackage{hyperref}

\usepackage{amsmath,amsfonts,bm}
\usepackage{upgreek}

\def\eqref#1{equation~\ref{#1}}

\def\1{\bm{1}}

\def\rw{{\textnormal{w}}}

\def\rvw{{\mathbf{w}}}
\def\rvx{{\mathbf{x}}}
\def\rvy{{\mathbf{y}}}

\def\rmH{{\mathbf{H}}}
\def\rmI{{\mathbf{I}}}
\def\rmJ{{\mathbf{J}}}

\def\rmM{{\mathbf{M}}}

\def\rmX{{\mathbf{X}}}

\def\vmu{{\bm{\mu}}}

\def\vf{{\bm{f}}}

\def\mSigma{{\bm{\Sigma}}}

\DeclareMathAlphabet{\mathsfit}{\encodingdefault}{\sfdefault}{m}{sl}
\SetMathAlphabet{\mathsfit}{bold}{\encodingdefault}{\sfdefault}{bx}{n}

\usepackage[capitalise]{cleveref}
\crefformat{equation}{(#2#1#3)}
\crefformat{appendix}{App. #2#1#3}
\usepackage{todonotes}
\usepackage{amsthm}
\usepackage{cancel}
\usepackage{enumitem}
\usepackage{placeins}
\usepackage{url}
\usepackage{glossaries}
\usepackage{nth}
\usepackage{empheq}
\usepackage{wrapfig}
\usepackage{nicefrac}

\usepackage[accepted]{icml2021}

\icmltitlerunning{Bayesian Deep Learning via Subnetwork Inference}

\begin{document}

\twocolumn[
\icmltitle{Bayesian Deep Learning via Subnetwork Inference}

\icmlsetsymbol{equal}{*}

\begin{icmlauthorlist}
\icmlauthor{Erik Daxberger}{cam,mpi}
\icmlauthor{Eric Nalisnick}{equal,uva}
\icmlauthor{James Urquhart Allingham}{equal,cam}
\icmlauthor{Javier Antor\'{a}n}{equal,cam}
\icmlauthor{Jos\'{e} Miguel Hern\'{a}ndez-Lobato}{cam,msr,ati}
\end{icmlauthorlist}

\icmlaffiliation{cam}{University of Cambridge}
\icmlaffiliation{mpi}{Max Planck Institute for Intelligent Systems, T\"{ubingen}}
\icmlaffiliation{uva}{University of Amsterdam}
\icmlaffiliation{msr}{Microsoft Research}
\icmlaffiliation{ati}{The Alan Turing Institute}

\icmlcorrespondingauthor{Erik Daxberger}{ead54@cam.ac.uk}

\icmlkeywords{Machine Learning, ICML}

\vskip 0.3in
]

\printAffiliationsAndNotice{\icmlEqualContribution} %

\begin{abstract}
    The Bayesian paradigm has the potential to solve core issues of deep neural networks such as poor calibration and data inefficiency.
    Alas, scaling Bayesian inference to large weight spaces often requires restrictive approximations.
    In this work, we show that it suffices to perform inference over a small subset of model weights in order to obtain accurate predictive posteriors. The other weights are kept as point estimates.
    This \emph{subnetwork inference framework} enables us to use expressive, otherwise intractable, posterior approximations over such subsets.
    In particular, we implement \emph{subnetwork linearized Laplace} as a simple, scalable Bayesian deep learning method:
    We first obtain a MAP estimate of all weights and then infer a full-covariance Gaussian posterior over a subnetwork using the linearized Laplace approximation.
    We propose a subnetwork selection strategy that aims to maximally preserve the model’s predictive uncertainty.
    Empirically, our approach compares favorably to ensembles and less expressive posterior approximations over full networks.
    Our proposed subnetwork (linearized) Laplace method is implemented within the \texttt{laplace} PyTorch library \cite{daxberger2021laplace} at \url{https://github.com/AlexImmer/Laplace}.
\end{abstract}
\section{Introduction}
A critical shortcoming of deep neural networks (NNs) is that they tend to be poorly calibrated and overconfident in their predictions, especially when there is a shift between the train and test data distributions \citep{nguyen2015,guo2017}.
To reliably inform decision making, NNs need to robustly quantify the \emph{uncertainty} in their predictions~\cite{bhatt2020uncertainty}. This is especially important for safety-critical applications such as healthcare or autonomous driving \citep{amodei2016}.

Bayesian modeling \citep{bishop2006pattern,ghahramani2015probabilistic} presents a principled way to capture uncertainty via the posterior distribution over model parameters.
Unfortunately,
exact posterior inference is intractable in NNs.
Despite recent successes in the field of Bayesian deep learning \citep{osawa2019,maddox2019,dusenberry2020efficient}, existing methods invoke unrealistic assumptions to scale to NNs with large numbers of weights.
This severely limits the expressiveness of the inferred posterior and thus deteriorates the quality of the induced uncertainty estimates \citep{ovadia2019,fort2019,foong2019expressiveness}.

\begin{figure*}[t]
    \centering
    \subfigure[Point Estimation]{\includegraphics[width=0.23\linewidth]{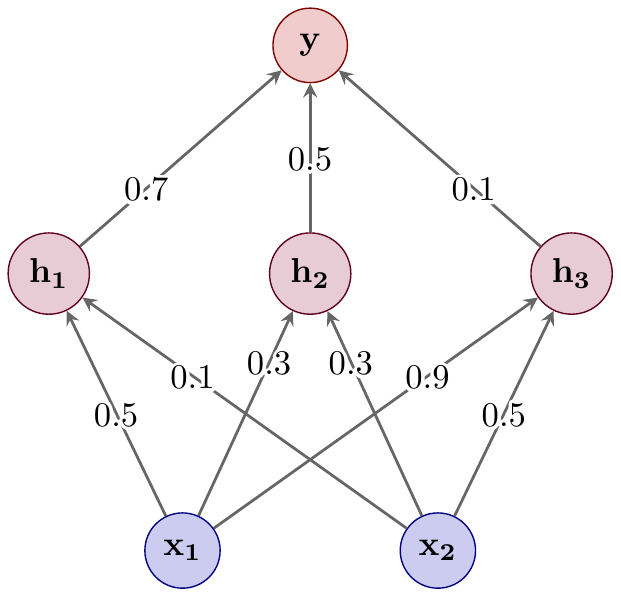}}
    \enskip
    \subfigure[Subnetwork Selection]{\includegraphics[width=0.23\linewidth]{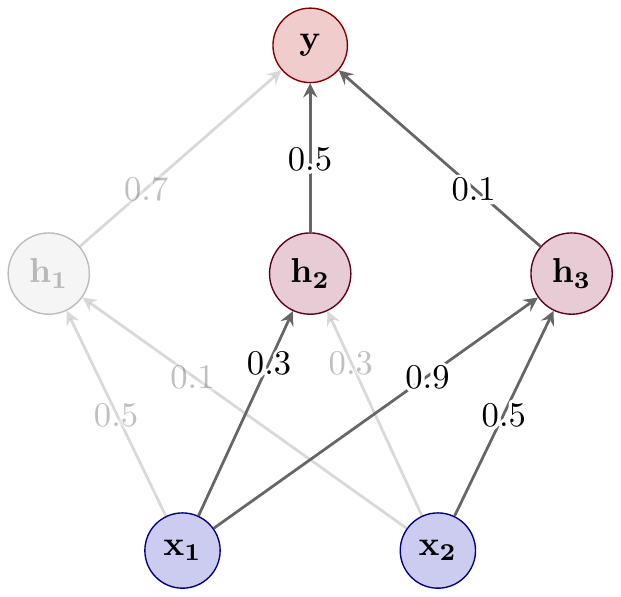}}
    \enskip
    \subfigure[Bayesian Inference]{\includegraphics[width=0.23\linewidth]{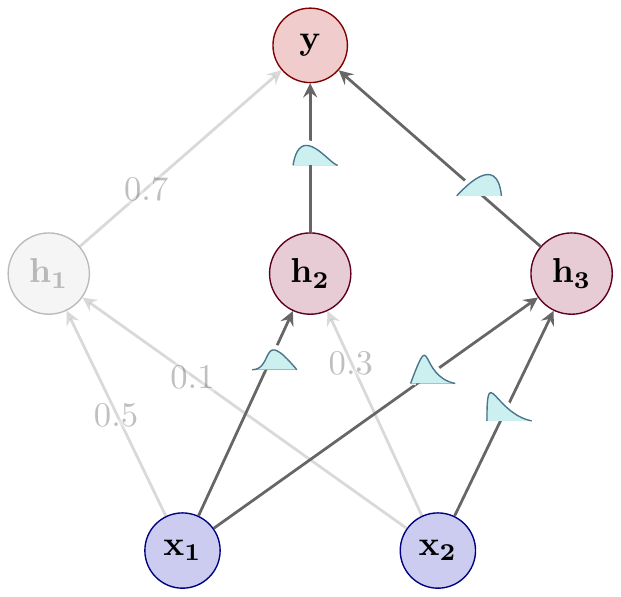}}
    \enskip
    \subfigure[Prediction]{\includegraphics[width=0.23\linewidth]{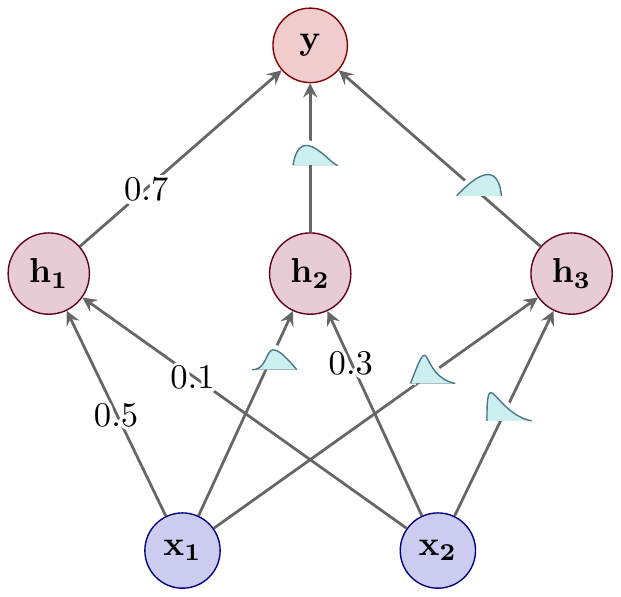}}
    \caption{Schematic illustration of our proposed approach. 
    (a) We train a neural network using standard techniques to obtain a point estimate of the weights.
    (b) We identify a small subset of the weights.
    (c) We estimate a posterior distribution over the selected subnetwork via Bayesian inference techniques.
    (d) We make predictions using the full network with a mix of Bayesian and deterministic weights.
    }
    \label{fig:schematic}
\end{figure*}

Perhaps these unrealistic inference approximations can be avoided. Due to the heavy overparameterization of NNs, their accuracy is well-preserved by a small subnetwork \citep{cheng2017survey}.
Moreover, doing inference over a low-dimensional subspace of the weights can result in accurate uncertainty quantification \cite{izmailov2019subspace}.
This prompts the following question:
\emph{Can a full NN's model uncertainty be well-preserved by a small subnetwork?}
In this work we demonstrate that the posterior predictive distribution of a full network \emph{can} be well represented by that of a subnetwork.
In particular, our contributions are as follows:
\begin{enumerate}[leftmargin=5mm]
    \item We propose \emph{subnetwork inference}, a general framework for scalable Bayesian deep learning in which inference is performed over only a \emph{small subset} of the NN weights, while all other weights are kept deterministic. This allows us to use \emph{expressive posterior approximations} that are typically intractable in large NNs. We present a concrete instantiation of this framework that first fits a MAP estimate of the full NN, and then uses the linearized Laplace approximation to infer a \emph{full-covariance Gaussian posterior} over a subnetwork (illustrated in \cref{fig:schematic}).
    \item We derive a subnetwork selection strategy based on the Wasserstein distance between the approximate posterior for the full network and the approximate posterior for the subnetwork. For scalability, we employ a diagonal approximation during subnetwork selection. Selecting a small subnetwork then allows us to  infer weight covariances. Empirically, we find that making approximations during subnetwork selection is much less harmful to the posterior predictive than making them during inference.
    \item We empirically evaluate our method on a range of benchmarks for \emph{uncertainty calibration} and \emph{robustness to distribution shift}. Our experiments demonstrate that expressive subnetwork inference can outperform popular Bayesian deep learning methods that do less expressive inference over the full NN as well as deep ensembles.
\end{enumerate}

\section{Subnetwork Posterior Approximation}
\label{sec:model}
Let $\rvw\in \mathbb{R}^D$ be the $D$-dimensional vector of all neural network weights (i.e.\ the concatenation and flattening of all layers' weight matrices).
Bayesian neural networks (BNNs) aim to capture \emph{model uncertainty}, i.e.~uncertainty about the choice of weights $\rvw$ arising due to multiple plausible explanations of the training data $\mathcal{D} \!= \!\{\rvy, \rmX\}$. Here, $\rvy \in \mathbb{R}^O$ is the output variable (e.g.\ classification label) and $\rmX \in \mathbb{R}^{N\times I}$ is the feature matrix.
First, a prior distribution $p(\rvw)$ is specified over the BNN's weights $\rvw$.
We then wish to infer their full \emph{posterior distribution}
\begin{equation}
    \label{eq:posterior}
    p(\rvw|\mathcal{D}) \ = \ p(\rvw | \rvy, \rmX) \ \propto \ p(\rvy | \rmX, \rvw) p(\rvw)\ .
\end{equation}
Finally, predictions for new data points $\rmX^*$ are made through marginalisation of the posterior:
\begin{equation}
    \label{eq:predictive}
    p(\rvy^*|\rmX^*, \mathcal{D}) = \int_{\rvw} p(\rvy^*|\rmX^*, \rvw) p(\rvw|\mathcal{D}) d\rvw\ .
\end{equation}
This \emph{posterior predictive} distribution translates uncertainty in weights to uncertainty in predictions.
Unfortunately, due to the non-linearity of NNs, it is intractable to infer the exact posterior distribution $p(\rvw| \mathcal{D})$.
It is even computationally challenging to faithfully \emph{approximate} the posterior due to the high dimensionality of $\rvw$.
Thus, crude posterior approximations such as complete factorization, i.e. $p(\rvw| \mathcal{D}) \approx \prod_{d=1}^{D}  q(\rw_d)$ where $\rw_d$ is the $d^\text{th}$ weight in $\rvw$, are commonly employed \citep{hernandez2015,blundell2015,khan2018,osawa2019}.
However, it has been shown that such an approximation suffers from severe pathologies \citep{foong2019expressiveness,foong2019between}.

In this work, we question the widespread implicit assumption that an expressive posterior approximation must include \emph{all} $D$ of the model weights. Instead, we try to perform inference only over a \emph{small subset} of $S \ll D$ of the weights.
The following arguments motivate this approach:
\begin{enumerate}[leftmargin=5mm]
    \item \textbf{Overparameterization:} \citet{maddox2020rethinking} have shown that, in the neighborhood of local optima, there are many directions that leave the NN's predictions unchanged.  Moreover, NNs can be heavily pruned without sacrificing test-set accuracy \citep{frankle2018the}.
    This suggests that the majority of a NN's predictive power can be isolated to a small subnetwork.
    \item  \textbf{Inference over submodels:} Previous work\footnote{See Section \ref{sec:related_work} for a more thorough discussion of related work.} 
    has provided evidence that inference can be effective even when not performed on the full parameter space.
    Examples include \citet{izmailov2019subspace} and \citet{snoek2015scalable} who perform inference over low-dimensional projections of the weights, and only the last layer of a NN, respectively.
\end{enumerate}  
We therefore combine these two ideas and make the following two-step approximation of the posterior in~\cref{eq:posterior}: 
\begin{align}
\label{eq:approximation1}
     p(\rvw| \mathcal{D}) \ \ &\approx \ \ p(\rvw_S | \mathcal{D}) \ \prod_{r} \delta(\rw_{r} - \widehat{\rw}_{r} )\\
\label{eq:approximation2}
     &\approx \ \ q(\rvw_S) \ \prod_{r} \delta(\rw_{r} - \widehat{\rw}_{r} ) \ \ = \ \ q_S(\rvw)\ .
\end{align}
The first approximation \cref{eq:approximation1} decomposes the full NN posterior $p(\rvw| \mathcal{D})$ into a posterior $p(\rvw_S |\mathcal{D})$ over the subnetwork $\rvw_S \in \mathbb{R}^S$ and Dirac delta functions $\delta(\rw_{r} - \widehat{\rw}_{r})$ over the $D-S$ remaining weights $\rw_r$ to keep them at fixed values $\widehat{\rw}_{r} \in \mathbb{R}$.
Since posterior inference over the subnetwork is still intractable, \cref{eq:approximation2} further approximates $p(\rvw_S | \mathcal{D})$ by $q(\rvw_S)$.
However, importantly, if the subnetwork is much smaller than the full network, we can afford to make $q(\rvw_S)$ \emph{more expressive} than would otherwise be possible. We hypothesize that being able to capture rich dependencies across the weights within the subnetwork will provide better results than crude approximations applied to the full set of weights.

\textbf{Relationship to Weight Pruning Methods.}
Note that the posterior approximation in \cref{eq:approximation2} can be viewed as \emph{pruning the variances} of the weights $\{\rw_r\}_r$ to zero.
This is in contrast to weight pruning methods \citep{cheng2017survey} that set the \emph{weights} themselves to zero.
I.e., weight pruning methods can be viewed as removing \emph{weights} to preserve the predictive \emph{mean} (i.e.\ to retain \emph{accuracy} close to the full model).
In contrast, subnetwork inference can be viewed as removing just the \emph{variances} of certain weights---while keeping their means---to preserve the predictive \emph{uncertainty} (e.g.\ to retain \emph{calibration} close to the full model).
Thus, they are complementary approaches.
Importantly, by not pruning weights, subnetwork inference retains the \emph{full predictive power} of the full NN to retain its predictive accuracy.

\section{Background: Linearized Laplace}\label{sec:lin_laplace}

In this work we satisfy \cref{eq:approximation2} by approximating the posterior distribution over the weights with the \emph{linearized Laplace approximation} \citep{mackay1992practical}. This is an inference technique that has recently been shown to perform strongly~\cite{foong2019between,immer2020improving} and can be applied \emph{post-hoc} to pre-trained models.
We now describe it in a general setting.
See \citet{daxberger2021laplace} for a more detailed description, review of recent advances, and software library for the Laplace approximation in deep learning.

We denote our NN function as $\vf: \mathbb{R}^{I} \to \mathbb{R}^{O}$.
We begin by defining a prior over our NN's weights, which we choose to be a fully factorised Gaussian $p(\rvw) = \mathcal{N}(\rvw; \textbf{0}, \lambda \rmI)$. We find a local optimum of the posterior, also known as a \textit{maximum a posteriori} (MAP) setting of the weights:
\begin{equation}\label{eq:MAP_estimate}
    \widehat{\rvw} = \operatorname{arg\, max}_{\rvw} \left[ \log p(\rvy | \rmX, \rvw) + \log p(\rvw) \right].
\end{equation}
The posterior is then approximated with a second order Taylor expansion around the MAP estimate:
\begin{gather}\label{eq:taylor_expansion}
    \log p(\rvw | \mathcal{D}) \approx \log p(\widehat{\rvw} | \mathcal{D})
    \!- \!\frac{1}{2} (\rvw - \widehat{\rvw})^\top\rmH (\rvw - \widehat{\rvw})
\end{gather}
where $\rmH \!\in\! \mathbb{R}^{D{\times}D}$ is the Hessian of the negative log-posterior density w.r.t.\ the network weights $\rvw$:
\begin{align}
    \rmH = N\cdot \mathbb{E}_{p(\mathcal{D})}\left[ - \partial^2 \log p(\rvy|\rmX, \rvw) / \partial \rvw^2 \right] + \lambda \rmI \,.
\end{align}
Thus, the approximate posterior takes the form of a full-covariance Gaussian with covariance matrix $\rmH^{-1}$:
\begin{equation}
\label{eq:laplace_posterior}
    p(\rvw | \mathcal{D}) \approx q(\rvw) = \mathcal{N}\left(\rvw; \widehat{\rvw}, \rmH^{-1}\right).
\end{equation}
In practise, the Hessian $\rmH$ is commonly replaced with the \emph{generalized Gauss-Newton matrix (GGN)} $\widetilde{\rmH} \in \mathbb{R}^{D \times D}$ \citep{martens2011learning,martens2014new,martens2016second}
\begin{equation}
\label{eq:ggn}
    \widetilde{\rmH} = \textstyle\sum_{n=1}^N \rmJ_n^\top \rmH_n \rmJ_n + \lambda \rmI\ .
\end{equation}
Here, $\rmJ_n = \partial \vf(\rvx_n, \rvw) / \partial \rvw \!\! \in \!\!\mathbb{R}^{O \times D}$ is the Jacobian of the model outputs $\vf(\rvx_n, \rvw) \!\!\in \!\! \mathbb{R}^O$ w.r.t.\ $\rvw$. $\rmH_n = - \partial^2 \log p(\rvy|\vf(\rvx_n, \rvw)) / \partial^2 \vf(\rvx_n, \rvw) \in \mathbb{R}^{O \times O}$ is the Hessian of the negative log-likelihood w.r.t.\ model outputs.

Interestingly, when using a Gaussian likelihood, the Gaussian with a GGN precision matrix corresponds to the \emph{true} posterior distribution when the NN is approximated with a first-order Taylor expansion around $\widehat{\rvw}$ \citep{khan2019approximate,immer2020improving}.
The \emph{locally linearized} function is
\begin{equation}
\label{eq:linearized}
\vf_{\text{lin}}(\rvx, \rvw) = \vf(\rvx, \widehat{\rvw}) + \widehat{\rmJ}(\rvx)(\rvw - \widehat{\rvw})
\end{equation}
where $\widehat{\rmJ}(\rvx) = \partial \vf(\rvx, \widehat{\rvw}) / \partial \widehat{\rvw} \in \mathbb{R}^{O \times D}$.
This turns the underlying probabilistic model from a BNN into a generalized linear model (GLM), where the Jacobian $\widehat{\rmJ}(\rvx)$ acts as a basis function expansion. Making predictions with the GLM $\vf_{\text{lin}}$ has been found to outperform the corresponding BNN $\vf$ with the GGN-Laplace posterior \cite{lawrence2001variational,foong2019between,immer2020improving}. Additionally, the equivalence between a GLM and a linearized BNN will help us to derive a subnetwork selection strategy in \cref{sec:subnet_selection}. 

The resulting posterior predictive distribution is
\begin{equation}
\label{eq:linearized_predictive}
    p(\rvy^*|\rvx^*, \mathcal{D}) = \!\int \! p(\rvy^*|\vf_{\text{lin}}(\rvx^*, \rvw)) p(\rvw | \mathcal{D})\,d\rvw\ .
\end{equation}
For regression, when using a Gaussian noise model $p(\rvy^*|\vf_{\text{lin}}(\rvx^*, \rvw)) \!= \mathcal{N}(\rvy^*; \vf_{\text{lin}}(\rvx^*, \rvw), \sigma^{2})$, our approximate distribution becomes exact $q(\rvw) = p(\rvw | \mathcal{D}) = \mathcal{N}(\rvw; \widehat{\rvw}, \widetilde{\rmH}^{-1})$.
We obtain the closed form predictive
\begin{align} \label{eq:regression_predictive}
    p(\rvy^*|\rvx^*, \mathcal{D}) = \mathcal{N}(\rvy^*; \vf(\rvx^*, \widehat{\rvw}), \Sigma(\rvx^*){+}\sigma^{2} \rmI)\ ,
\end{align}
where $\Sigma(\rvx^*) \! = \! \widehat{\rmJ}(\rvx^*)^\top\widetilde{\rmH}^{-1}\widehat{\rmJ}(\rvx^*)$.
For classification with a categorical likelihood $p(\rvy^*|\vf_{\text{lin}}(\rvx^*, \rvw)) = \text{Cat}(\rvy^*; \phi(\vf_{\text{lin}}(\rvx^*, \rvw))$, the posterior is strictly convex. This makes our Gaussian a faithful approximation. Here, $\phi(\cdot)$ refers to the softmax function.
The predictive integral has no analytical solution. Instead we leverage the probit approximation \citep{gibbs1998bayesian,bishop2006pattern}: 
\begin{align} \label{eq:classification_predictive}
\hspace{-3mm}
    p(\rvy^* | \rvx^*, \mathcal{D}) \!\approx\! \text{Cat}\!\left(\!\rvy^*; \phi\!\left(\frac{ \vf(\rvx^*, \widehat{\rvw})}{\sqrt{1{+}\frac{\pi}{8} \text{diag}( \Sigma(\rvx^*))}} \right)\!\right).
\end{align}
These closed-form expressions are attractive since they result in the predictive mean and classification boundaries being exactly equal to those of the MAP estimated NN.

Unfortunately, storing the full $D \times D$ covariance matrix over the weight space of a modern NN (i.e.\ with very large $D$) is computationally intractable.
There have been efforts to develop cheaper approximations to this object, such as only storing diagonal \citep{denker1990transforming} or block diagonal \citep{ritter2018a,immer2020improving} entries, but these come at the cost of reduced predictive performance.

\section{Linearized Laplace Subnetwork Inference}
\label{sec:subnetwork_inference}
We outline the following procedure for scaling the linearized Laplace approximation to large neural network models within the framework of subnetwork inference.

\paragraph{Step \#1: Point Estimation, \cref{fig:schematic}~(a).}
Train a neural network to obtain a point estimate of the weights, denoted $\widehat{\rvw}$.
This can be done using stochastic gradient-based optimization methods \citep{goodfellow2016}.
Alternatively, we could make use of a pre-trained model.

\paragraph{Step \#2: Subnetwork Selection, \cref{fig:schematic} (b).} 
Identify a small subnetwork $\rvw_S \in \mathbb{R}^S, S \ll D$.  Ideally, we would like to find the subnetwork which produces a predictive posterior `closest' to the full-network's predictive distribution. Regrettably, reasoning in the space of functions directly is challenging \citep{burt2020understanding}. Instead, in  \cref{sec:subnet_selection}, we describe a strategy that minimizes the Wasserstein distance between the sub- and full-network's weight posteriors.

\paragraph{Step \#3: Bayesian Inference, \cref{fig:schematic}~(c).}

Use the GGN-Laplace approximation to infer a full-covariance Gaussian posterior over the subnetwork's weights $\rvw_S \in \mathbb{R}^S$:
\begin{equation}
\label{eq:subnetwork_laplace_posterior}
    p(\rvw_S | \mathcal{D}) \approx q(\rvw_S) = \mathcal{N}(\rvw_S; \widehat{\rvw}_S, \widetilde{\rmH}_S^{-1})
\end{equation}
where $\widetilde{\rmH}_S \in \mathbb{R}^{S{\times}S}$ is the GGN w.r.t.\ the weights $\rvw_S$:
\begin{equation}
\label{eq:subnetwork_ggn}
    \widetilde{\rmH}_{S} = \textstyle\sum_{n=1}^N \rmJ_{Sn}^\top \rmH_n \rmJ_{Sn} + \lambda_{S} \rmI\ .
\end{equation}
Here, $\rmJ_{Sn} = \partial \vf(\rvx_n, \rvw_{S}) / \partial \rvw_{S} \!\! \in \!\!\mathbb{R}^{O \times S}$ is the Jacobian w.r.t.\ $\rvw_{S}$. $\rmH_n$ is defined as in \cref{sec:model}. In order to best preserve the magnitude of the predictive variance, we update our prior precision to be $\lambda_{S} = \lambda \! \cdot \! \nicefrac{S}{D}$ (see \cref{app:prior_update} for more details).
All weights not belonging to the chosen subnetwork are fixed at their MAP values.
Note that this whole procedure (i.e.\ Steps \#1-\#3) is a perfectly valid mixed inference strategy: We perform full Laplace inference over the selected subnetwork and MAP inference over all remaining weights.
The resulting approximate posterior \cref{eq:approximation2} is
\begin{equation}
\label{eq:approximate_posterior_linearized}
    q_{S}(\rvw) \overset{(\ref{eq:subnetwork_laplace_posterior})}{=} \mathcal{N}(\rvw_{S}; \widehat{\rvw}_{S}, \widetilde{\rmH}_S^{-1}) \textstyle\prod_{r} \delta(\rw_r - \widehat{\rw}_r )\ .
\end{equation}
Given a sufficiently small subnetwork $\rvw_{S}$, it is feasible to store and invert $\widetilde{\rmH}_{S}$.
In particular, naively storing and inverting the \emph{full} GGN $\widetilde{\rmH}$ scales as $\mathcal{O}(D^2)$ and $\mathcal{O}(D^3)$, respectively.
Using the subnetwork GGN $\widetilde{\rmH}_{S}$ instead reduces this burden to $\mathcal{O}(S^2)$ and $\mathcal{O}(S^3)$, respectively. In our experiments, $S\,{\ll}\,D$ with our subnetworks representing less that 1\% of the total weights.
Note that quadratic/cubic scaling in $S$ is unavoidable if we are to capture weight correlations.

\paragraph{Step \#4: Prediction, \cref{fig:schematic}~(d).
}
Perform a local linearization of the NN (see \cref{sec:lin_laplace}) while fixing $\rvw_{r}$ to $\widehat{\rvw}_{r}$:
\begin{equation}
\label{eq:subnetwork_linearized}
    \vf_{\text{lin}}(\rvx, \rvw_{S}) = \vf(\rvx, \widehat{\rvw}) + \widehat{\rmJ}_{S}(\rvx)(\rvw_{S} - \widehat{\rvw}_{S})\ ,
\end{equation}
where $\widehat{\rmJ}_{S}(\rvx) = \partial \vf(\rvx, \widehat{\rvw}_{S}) / \partial \widehat{\rvw}_{S} \! \in \! \mathbb{R}^{O \times S}$.
Following \cref{eq:regression_predictive} and \cref{eq:classification_predictive},
the corresponding predictive distributions are 
\begin{align} \label{eq:regression_predictive_subnet}
    p(\rvy^*|\rvx^*, \mathcal{D}) = \mathcal{N}(\rvy^*; \vf(\rvx^*, \widehat{\rvw}), \textcolor{OrangeRed}{\Sigma_{S}(\rvx^*)}{+}\sigma^{2} \rmI)
\end{align}
for regression and
\begin{align} \label{eq:classification_predictive_subnet}
    p(\rvy^* | \rvx^*, \mathcal{D}) \approx  \text{softmax}\left(\frac{ \vf(\rvx^*, \widehat{\rvw})}{\sqrt{1{+}\frac{\pi}{8} \text{diag}( \textcolor{OrangeRed}{\Sigma_{S}(\rvx^*)})}} \right)
\end{align}
for classification, where $\Sigma(\rvx^{*})$ in \cref{eq:regression_predictive} and \cref{eq:classification_predictive} is substituted with $\textcolor{OrangeRed}{\Sigma_{S}(\rvx^{*})} = \widehat{\rmJ}_{S}(\rvx^*)^{T}\widetilde{\rmH}_{S}^{-1}\widehat{\rmJ}_{S}(\rvx^*)$.

\begin{figure*}[t]
    \centering
    \includegraphics[width=0.95\textwidth]{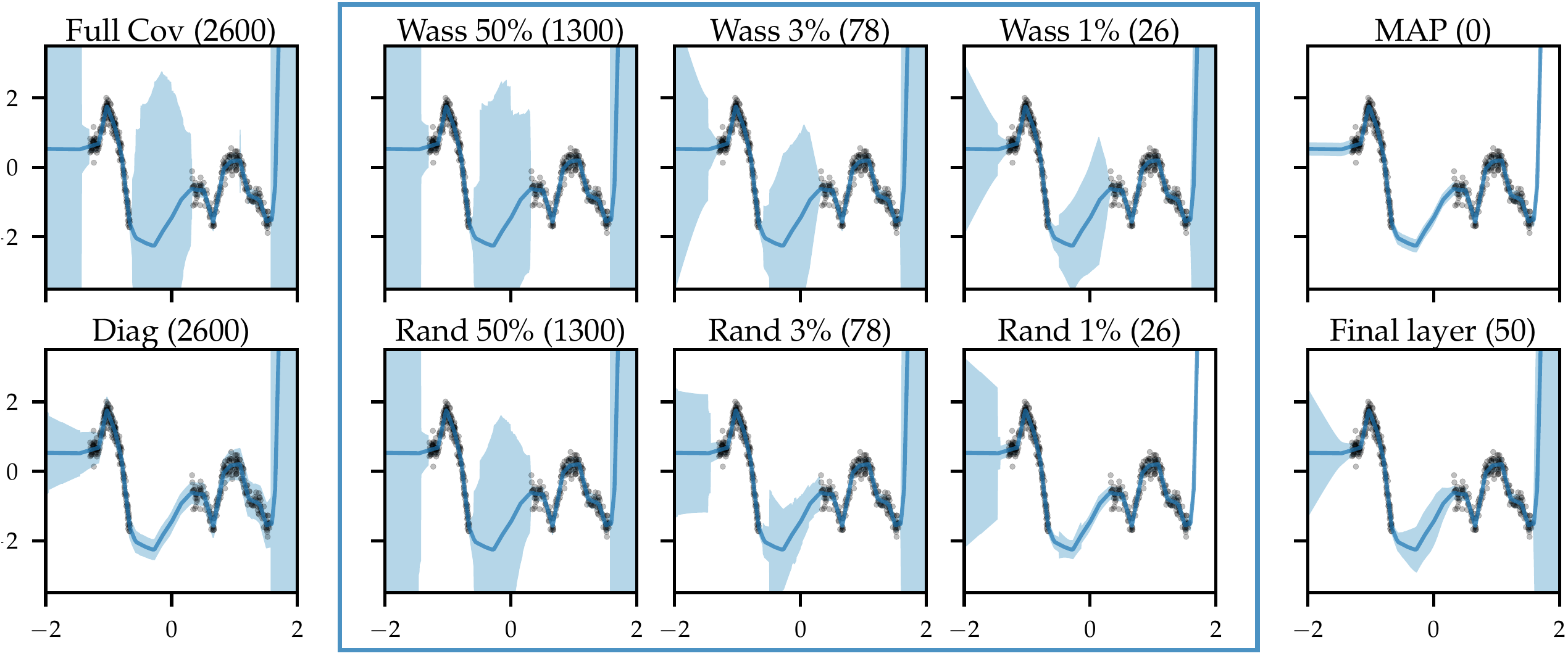}
    \caption{Predictive distributions (mean $\pm$ std) for 1D regression. The numbers in parentheses denote the number of parameters over which inference was done (out of 2600 in total). The blue box highlights subnetwork inference using Wasserstein (top) and random (bottom) subnetwork selection. Wasserstein subnetwork inference maintains richer predictive uncertainties at smaller parameter counts.}
    \label{fig:toy_exp}
\end{figure*}

\section{Subnetwork Selection}
\label{sec:subnet_selection}

Ideally, we would like to choose a subnetwork such that the induced predictive posterior distribution is as close as possible to the predictive posterior provided by inference over the full network \cref{eq:linearized_predictive}. This discrepancy between stochastic processes is often quantified through the functional Kullback-Leibler (KL) divergence \citep{sun2019functional,burt2020understanding}:
\begin{align}\label{eq:functional_KL}
    \sup_{n \in {\mathbb{N}}, \rmX^{*} \in \mathcal{X}^{n}} \!\!\!D_{\text{KL}}(p_{S}(\rvy^*|\rmX^{*}, \mathcal{D})\,||\,p(\rvy^*|\rmX^{*}, \mathcal{D})) ,
\end{align}
where $p_{S}$ denotes the subnetwork predictive posterior and $\mathcal{X}^{n}$ denotes a finite measurement set of $n$ elements. 
Regrettably, reasoning directly in function space is a difficult task \citep{nalisnick2018learning, pearce2019expressive, sun2019functional, antoran2020depth, nalisnick2020predictive, burt2020understanding}. Instead we focus our attention on weight space.

In weight space, our aim is to minimise the discrepancy between the exact posterior over the full network \cref{eq:posterior} and the subnetwork approximate posterior \cref{eq:approximation2}. This provides two challenges. Firstly, computing the exact posterior distribution remains intractable. Secondly, common discrepancies, like the KL divergence or the Hellinger distance, are not well defined for the Dirac delta distributions found in \cref{eq:approximation2}.

To solve the first issue, we again resort to local linearization, introduced in \cref{sec:lin_laplace}. The true posterior for the linearized model is Gaussian or approximately Gaussian\footnote{When not making predictions with the linearized model, the Gaussian posterior would represent a crude approximation.}:
\begin{equation}
\label{eq:lin_laplace_posterior}
    p(\rvw | \mathcal{D}) \simeq \mathcal{N}(\rvw; \widehat{\rvw}, \widetilde{\rmH}^{-1})\ .
\end{equation}
We solve the second issue by choosing the squared 2-Wasserstein distance, which is well defined for distributions with disjoint support.
For the case of a full covariance Gaussian \cref{eq:lin_laplace_posterior} and a product of a full covariance Gaussian with Dirac deltas \cref{eq:approximate_posterior_linearized}, this metric takes the following form:
\begin{align}
    &W_2(p(\rvw | \mathcal{D}), q_{S}(\rvw))^2 \label{eq:wass2squared} \\ \notag
    &= \text{Tr}\left(\widetilde{\rmH}^{-1} + \widetilde{\rmH}_{S+}^{-1} - 2 \left(\widetilde{\rmH}_{S+}^{-1/2}\widetilde{\rmH}^{-1}\widetilde{\rmH}_{S+}^{-1/2}\right)^{1/2} \right)\ , 
\end{align}
where the covariance matrix $\widetilde{\rmH}_{S+}^{-1}$ is equal to $\widetilde{\rmH}_{S}^{-1}$ padded with zeros at the positions corresponding to $\rw_r$, matching the shape of $\widetilde{\rmH}^{-1}$. See \cref{sec:proofs} for details.

Finding the subset $\rvw_S \in \mathbb{R}^S$ of size $S$ that minimizes \cref{eq:wass2squared} would be combinatorially difficult, as the contribution of each weight depends on every other weight. 
To address this issue, we make an independence assumption among weights, resulting in the simplified objective
\begin{align}
\label{eq:wass2squared_indep}
    W_2(p(\rvw | \mathcal{D}), q_{S}(\rvw))^2 
    \approx \sum_{d=1}^{D} \sigma^2_d(1 - m_d)\ , 
\end{align}
where $\sigma^2_d$ is the \emph{marginal variance} of the $d^{\text{th}}$ weight, and $m_d = 1$ if $\rw_d\,\in\,\rvw_S$ and 0 otherwise (see \cref{sec:proofs}). 
The objective \cref{eq:wass2squared_indep} is trivially minimized by a subnetwork containing the $S$ weights with highest variances.
This is related to common magnitude-based weight pruning methods \citep{cheng2017survey}.
The main difference is that our selection strategy involves weight \emph{variances} rather than \emph{magnitudes} as we target predictive uncertainty rather than accuracy.

In practice, even computing the marginal variances (i.e.\ the diagonal of $\widetilde{\rmH}^{-1}$) is intractable, as it requires storing and inverting the GGN $\widetilde{\rmH}$.
However, we can approximate posterior marginal variances with the diagonal Laplace approximation $\text{diag}(\widetilde{\rmH}^{-1}) \! \approx \! \text{diag}(\widetilde{\rmH})^{-1}$ \citep{denker1990transforming,kirkpatrick2017overcomign}, diagonal SWAG \citep{maddox2019}, or even mean-field variational inference \citep{blundell2015,osawa2019}. In this work we rely on the former two, as the the latter involves larger overhead.

It may seem that we have resorted to the poorly performing diagonal assumptions that we sought to avoid in the first place~\cite{ovadia2019,foong2019expressiveness,ashukha2020pitfalls}. However, there is a key difference. We make the diagonal assumption during \emph{subnetwork selection} rather than \emph{inference}; we do full covariance inference over $\rvw_S$. In \cref{sec:experiments}, we provide evidence that making a diagonal assumption during subnetwork selection is reasonable by showing that 1) it is substantially less harmful to predictive performance than making the same assumption during inference, and 2) it outperforms random subnetwork selection.

\section{Experiments}
\label{sec:experiments}
We empirically assess the effectiveness of subnetwork inference compared to methods that do less expressive inference over the full network as well as state-of-the-art methods for uncertainty quantification in deep learning.
We consider three benchmark settings: 1) small-scale toy regression, 2) medium-scale tabular regression, and 3) image classification with ResNet-18.
Further experimental results and setup details are presented in \cref{sec:additional_results} and \cref{sec:exp_setup}, respectively.
\begin{figure*}[ht]
    \centering
    \includegraphics[width=\textwidth]{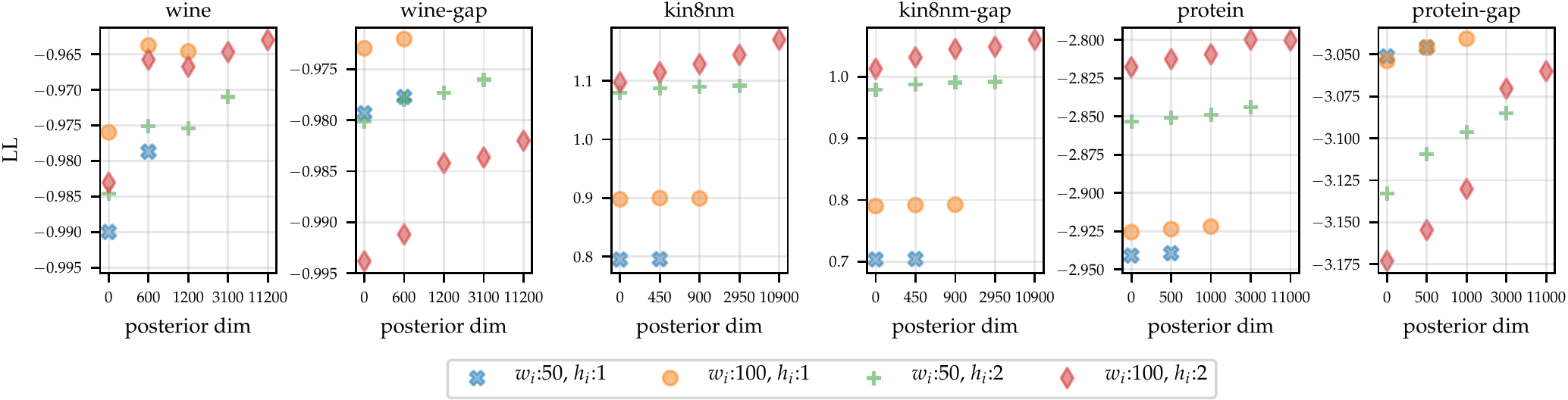}
    \caption{Mean test log-likelihood values obtained on UCI datasets across all splits. Different markers indicate models with different numbers of weights. The horizontal axis indicates the number of weights over which full covariance inference is performed. 0 corresponds to MAP parameter estimation, and the rightmost setting for each marker corresponds to full network inference.}
    \label{fig:UCI_exp}
\end{figure*}

\subsection{How does Subnetwork Inference preserve Posterior Predictive Uncertainty?}
\label{sec:toy_1d}
We first assess how the predictive distribution of a full-covariance Gaussian posterior over a selected subnetwork qualitatively compares to that obtained from 1) a full-covariance Gaussian over the \emph{full} network (\emph{Full Cov}), 2) a \emph{factorised} Gaussian posterior over the full network (\emph{Diag}), 3) a full-covariance Gaussian over only the (\emph{Final layer}) of the network \citep{snoek2015scalable}, and 4) a point estimate (\emph{MAP}).
For subnetwork inference, we consider both Wasserstein (\emph{Wass}) (as described in \cref{sec:subnet_selection}) and uniform random subnetwork selection (\emph{Rand}) to obtain subnetworks that comprise of only 50\%, 3\% and 1\% of the model parameters.
For this toy example, it is tractable to compute exact posterior marginal variances to guide subnetwork selection.

Our NN consists of 2 ReLU hidden layers with 50 hidden units each. We employ a homoscedastic Gaussian likelihood function where the noise variance is optimised with maximum likelihood.
We use GGN-Laplace inference over network weights (not biases) in combination with the linearized predictive distribution in \cref{eq:regression_predictive_subnet}.
Thus, all approaches considered share their predictive mean, allowing better comparison of their uncertainty estimates. We set the full network prior precision to $\lambda = 3$ (a value which we find to work well empirically) and set $\lambda_{S} = \lambda \cdot \nicefrac{S}{D}$.

We use a synthetic 1D regression task with two separated clusters of inputs \citep{antoran2020depth}, allowing us to probe for `in-between' uncertainty \citep{foong2019between}.
Results are shown in \cref{fig:toy_exp}.
Subnetwork inference preserves more of the uncertainty of full network inference than diagonal Gaussian or final layer inference while doing inference over fewer weights. By capturing weight correlations, subnetwork inference retains uncertainty in between clusters of data. This is true for both random and Wasserstein subnetwork selection. However, the latter preserves more uncertainty with smaller subnetworks.
Finally, the strong superiority to diagonal Laplace shows that making a diagonal assumption for subnetwork selection but then using a full-covariance Gaussian for inference (as we do) performs significantly better than making a diagonal assumption for the inferred posterior directly (cf.\ \cref{sec:subnet_selection}).
These results suggest that \textbf{expressive inference over a carefully selected subnetwork retains more predictive uncertainty than crude approximations over the full network}.

\begin{figure*}[t]
    \centering
    \includegraphics[width=0.95\textwidth]{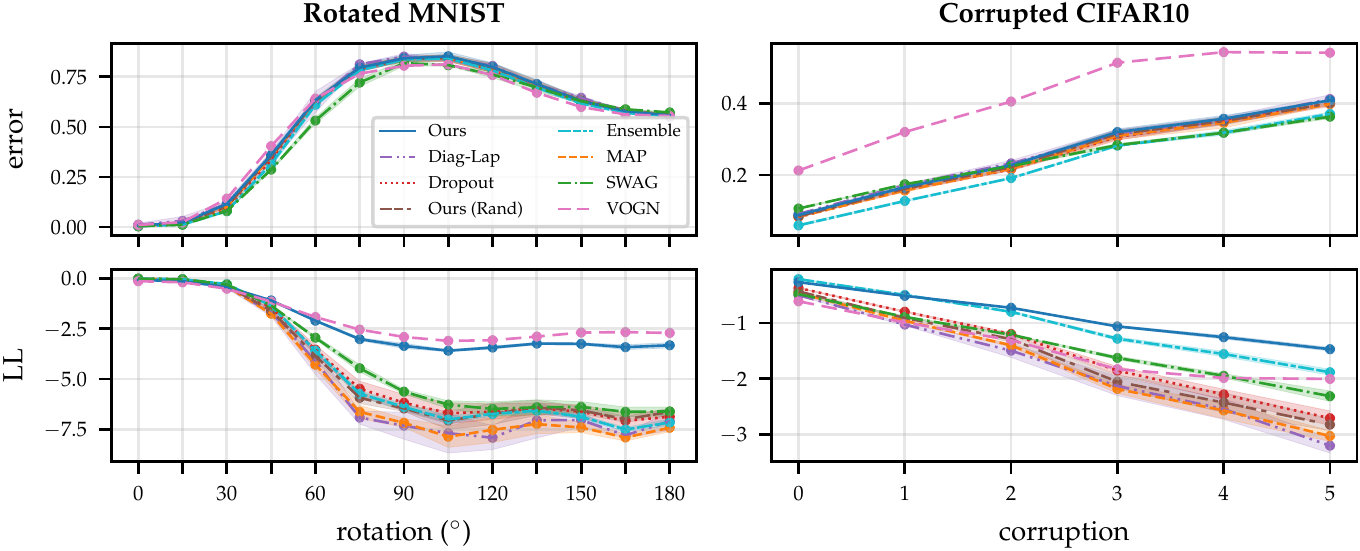}
    \caption{Results on the rotated MNIST (left) and the corrupted CIFAR (right) benchmarks, showing the mean $\pm$ std of the error (top) and log-likelihood (bottom) across three different seeds. Subnetwork inference retains better uncertainty calibration and robustness to distribution shift than point-estimated networks and other Bayesian deep learning approaches. See \cref{sec:additional_results} for ECE and Brier score results.}
    \label{fig:img_res}
\end{figure*}

\subsection{Subnetwork Inference in Large Models vs Full Inference over Small Models}\label{sec:regression_UCI}
Secondly, we study how subnetwork inference in larger NNs compares to full network inference in smaller ones.
We explore this by considering 4 fully connected NNs of increasing size. These have numbers of hidden layers $h_{d}{=}\{1, 2\}$ and hidden layer widths $w_{d}{=}\{50, 100\}$. For a dataset with input dimension $i_{d}$, the number of weights is given by $D{=}(i_{d}{+}1) w_{d}{+} (h_{d}{-}1)w_{d}^{2}$. Our 2 hidden layer, 100 hidden unit NNs have a weight count of the order $10^4$. Full covariance inference in these NNs borders the limit of computational tractability on commercial hardware.
We first obtain a MAP estimate of each NN’s weights and our homoscedastic likelihood function's noise variance. 
We then perform full network GGN-Laplace inference for each NN. We also use our proposed Wassertein rule to prune every NN’s weight variances such that the number of variances that remain matches the size of every smaller NN under consideration. We employ the diagonal Laplace approximation to cheaply estimate posterior marginal variances for subnetwork selection.
We employ the linearization in \cref{eq:regression_predictive} and \cref{eq:regression_predictive_subnet} to compute predictive distributions. Consequently, NNs with the same number of weights make the same mean predictions. Increasing the number of weight variances considered will thus only increase predictive uncertainty. 

We employ 3 tabular datasets of increasing size (input dimensionality, n. points): wine (11, 1439), kin8nm (8, 7373) and protein (9, 41157).
We consider their standard train-test splits \citep{hernandez2015} and their gap variants \citep{foong2019between}, designed to test for out-of-distribution uncertainty.  Details are provided in \cref{app:data}.
For each split, we set aside 15\% of the train data as a validation set. We use these for early stopping when finding MAP estimates and for selecting the weights’ prior precision.
We keep other hyperparameters fixed across all models and datasets. Results are shown in \cref{fig:UCI_exp}. 

We present mean test log-likelihood (LL) values, as these take into account both accuracy and uncertainty. 
Larger ($w_{d}=100, h_{d}=2$) models tend to perform best when combined with full network inference, although Wine-gap and Protein-gap are exceptions. Interestingly, these larger models are still best when we perform inference over subnetworks of the size of smaller models.
We conjecture this is due to an abundance of degenerate directions (i.e.\ weights) in the weight posterior NN models \citep{maddox2020rethinking}. Full network inference in small models captures information about both useful and non-useful weights. In larger models, our subnetwork selection strategy allows us to dedicate a larger proportion of our resources to modelling informative weight variances and covariances.
In 3 out of 6 datasets, we find abrupt increases in LL as we increase the number of weights over which we perform inference, followed by a plateau. Such plateaus might be explained by most of the informative weight variances having already been accounted for. 
Considering that the cost of computing the GGN dominates that of NN training, these results suggest that, \textbf{given the same amount of compute, it is better to perform subnetwork inference in larger models than full network inference in small ones.}
\begin{figure*}[ht]
    \centering
    \subfigure[Rotated MNIST]{\includegraphics[width=0.36\textwidth]{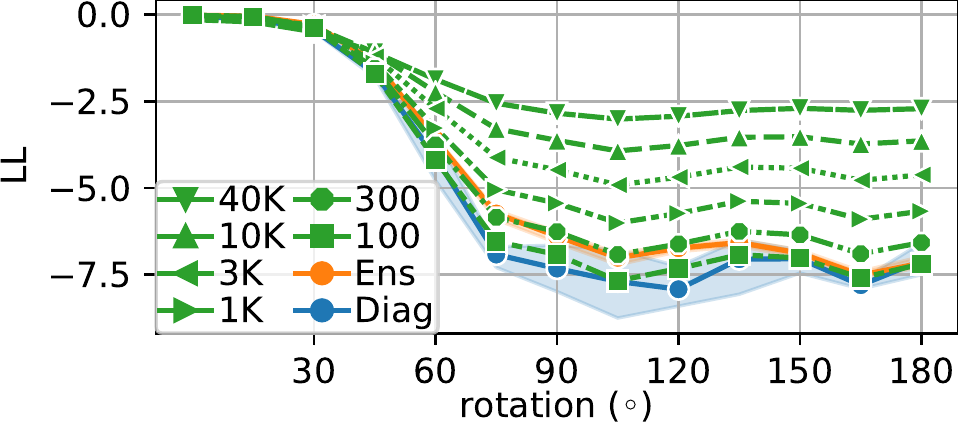}}%
    \enskip%
    \subfigure[Corrupted CIFAR10]{\includegraphics[width=0.36\textwidth]{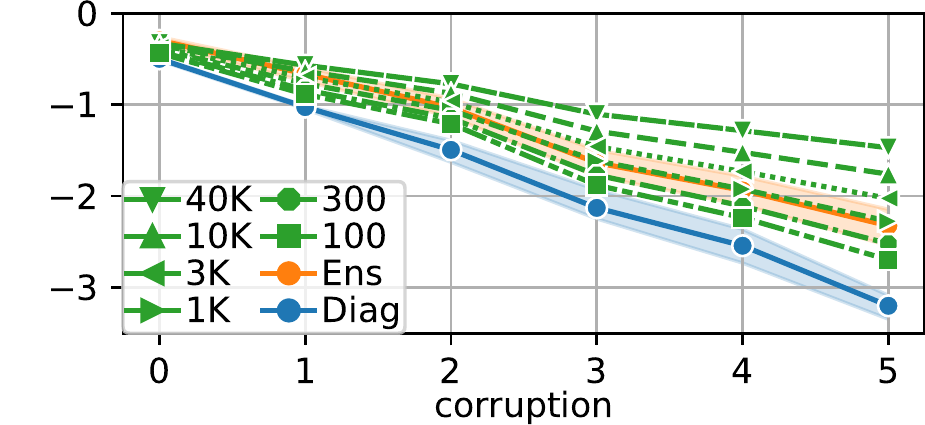}}%
    \enskip%
    \subfigure[Memory Footprints]{
        \begin{tabular}[b]{rr}
            \toprule
            \textbf{Subnet Size} & \textbf{Memory}\\
            \midrule
            11.2M (100\%) & 500\textbf{TB}\\
            \midrule
            40K (0.36\%) & 6.4\textbf{GB}\\
            1K (0.01\%) & 4.0\textbf{MB}\\
            100 (0.001\%) & 40\textbf{KB}\\
            \bottomrule
        \end{tabular}%
    }
    \caption{Log-likelihoods of our method with subnetwork sizes between 100-40K using ResNet-18 on rotated MNIST (left) and corrupted CIFAR10 (middle), vs.\ \textbf{Ens}embles and \textbf{Diag}onal Laplace, and respective covariance matrix memory footprints (right).
    For all subnetwork sizes, we use the same hyperparameters as in \cref{sec:img_exp} (i.e.\ no individual tuning per size).
    Performance degrades smoothly with subnetwork size, but our method retains strong calibration even with very small subnetworks (requiring only marginal extra memory).}
    \label{fig:subnet_size}
\end{figure*}

\subsection{Image Classification under Distribution Shift} 
\label{sec:img_exp}
We now assess the robustness of large convolutional neural networks with subnetwork inference to distribution shift on image classification tasks compared to the following baselines: 
point-estimated networks (MAP), Bayesian deep learning methods that do less expressive inference over the full network: MC Dropout \citep{gal2016}, diagonal Laplace, VOGN \citep{osawa2019} (all of which assume factorisation of the weight posterior), and SWAG \citep{maddox2019} (which assumes a diagonal plus low-rank posterior). We also benchmark deep ensembles \citep{lakshminarayanan2017}. The latter is considered state-of-the-art for uncertainty quantification in deep learning \citep{ovadia2019,ashukha2020pitfalls}. We use ensembles of 5 NNs, as suggested by \citep{ovadia2019}, and 16 samples for MC Dropout, diagonal Laplace and SWAG.
We use a Dropout probability of $0.1$ and a prior precision of $\lambda = 4 \times 10^4$ for diagonal Laplace, found via grid search.
We apply all approaches to ResNet-18 \citep{he2016deep}, which is composed of an input convolutional block, 8 residual blocks and a linear layer, for a total of 11,168,000 parameters.

For subnetwork inference, we compute the linearized predictive distribution in \cref{eq:classification_predictive_subnet}.
We use Wasserstein subnetwork selection to retain only $0.38\%$ of the weights, yielding a subnetwork with only 42,438 weights. This is the largest subnetwork for which we can tractably compute a full covariance matrix. Its size is $42,438^2 \times 4 ~\text{Bytes} \approx 7.2~\text{GB}$.
We use diagonal SWAG \citep{maddox2019} to estimate the marginal weight variances needed for subnetwork selection.
We tried diagonal Laplace but found that the selected weights where those where the Jacobian of the NN evaluated at the train points was always zero (i.e.\ dead ReLUs). The posterior variance of these weights is large as it matches the prior. However, these weights have little effect on the NN function. SWAG does not suffer from this problem as it disregards weights with zero training gradients.
We use a prior precision of $\lambda = 500$, found via grid search.

To assess to importance of principled subnetwork selection, we also consider the baseline where we select the subnetwork uniformly at random (called \emph{Ours (Rand)}).
We perform the following two experiments, with results in \cref{fig:img_res}.

\textbf{Rotated MNIST:} Following \cite{ovadia2019, antoran2020depth}, we train all methods on MNIST and evaluate their predictive distributions on increasingly rotated digits.
While all methods perform well on the original MNIST test set, their accuracy degrades quickly for rotations larger than 30 degrees. In terms of LL, ensembles perform best out of our baselines. Subnetwork inference obtains significantly larger LL values than almost all baselines, including ensembles. The only exception is VOGN, which achieves slightly better performance. It was also observed in \cite{ovadia2019} that mean-field variational inference (which VOGN is an instance of) is very strong on MNIST, but its performance deteriorates on larger datasets. Subnetwork inference makes accurate predictions in-distribution while assigning higher uncertainty than the baselines to out-of-distribution points.

\textbf{Corrupted CIFAR:}
Again following \cite{ovadia2019,antoran2020depth}, we train on CIFAR10 and evaluate on data subject to 16 different corruptions with 5 levels of intensity each \citep{hendrycks2019benchmarking}. Our approach matches a MAP estimated network in terms of predictive error as local linearization makes their predictions the same. Ensembles and SWAG are the most accurate. 
Even so, subnetwork inference differentiates itself by being the least overconfident, outperforming all baselines in terms of log-likelihood at all corruption levels. Here, VOGN performs rather badly; while this might appear to contrast its strong performance on the MNIST benchmark, the behaviour that mean-field VI performs well on MNIST but poorly on larger datasets was also observed in \cite{ovadia2019}.

Furthermore, on both benchmarks, we find that randomly selecting the subnetwork performs substantially worse than using our more sophisticated Wasserstein subnetwork selection strategy. This highlights the importance of the way the subnetwork is selected.
Overall, these results suggest that \textbf{subnetwork inference results in better uncertainty calibration and robustness to distribution shift than other popular uncertainty quantification approaches}.

\paragraph{What about smaller subnetworks?}
One might wonder if a subnetwork of $\sim$40K weights is actually necessary.
In \cref{fig:subnet_size}, we show that one can also retain strong calibration with significantly smaller subnetworks.
Full covariance inference in a ResNet-18 would require storing $\sim$11.2M$^{2}$ params ($\sim$500TB).
Subnet inference reduces the cost (on top of MAP) to as little as 1K$^2$ params ($\sim$4.0MB) while remaining competitive with deep ensembles.
This suggests that subnetwork inference can allow otherwise intractable inference methods to be applied to even larger NNs.
\section{Scope and Limitations}

\textbf{Jacobian computation in multi-output models} remains challenging. With reverse mode automatic differentiation used in most deep learning frameworks, it requires as many backward passes as there are model outputs. This prevents using linearized Laplace in settings like semantic segmentation \citep{Liu2019semantic} or classification with large numbers of classes \citep{imagenet}.
Note that this issue applies to the linearized Laplace method and that other inference methods, without this limitation, could be used in our framework.

\textbf{The choice of prior precision $\mathbf{\lambda}$} determines the performance of the Laplace approximation to a large degree. Our proposed scheme to update $\lambda$ for subnetworks relies on having a sensible parameter setting for the full network. Since inference in the full network is often intractable, currently the best approach for choosing $\lambda$ is cross validation using the subnetwork approximation directly.

\textbf{The space requirements for the Hessian} limit the maximum number of subnetwork weights.
For example, storing a Hessian for 40K weights requires around 6.4GB of memory.
For very large models, like modern transformers, tractable subnetworks would represent a vanishingly small proportion of the weights.  
While we demonstrated that strong performance does not necessarily require large subnetworks (see \cref{fig:subnet_size}), finding better subnetwork selection strategies remains a key direction for future research. 

\section{Related Work}
\label{sec:related_work}
\textbf{Bayesian Deep Learning.}
There have been significant efforts to characterise the posterior distribution over NN weights $p(\rvw|\mathcal{D})$. 
To this day, Hamiltonian Monte Carlo \citep{neil_thesis} remains the gold standard for approximate inference in BNNs \citep{izmailov2021bayesian}.
Although asymptotically unbiased, sampling based approaches are difficult to scale to large datasets \citep{betancourt2015fundamental}. As a result, approaches which find the best surrogate posterior among an approximating family (most often Gaussians) have gained popularity. The first of these was the Laplace approximation, introduced by \citet{mackay1992practical}, who also proposed approximating the predictive posterior with that of the linearised model \citep{khan2019approximate,immer2020improving};
see \citet{daxberger2021laplace} for a review of recent advances to use the Laplace approximation in deep learning.
The popularisation of larger NN models has made surrogate distributions that capture correlations between weights computationally intractable. Thus, most modern methods make use of the mean field assumption \citep{blundell2015,hernandez2015,gal2016,mishkin2018,osawa2019}. This comes at the cost of limited expressivity \citep{foong2019expressiveness} and empirical under-performance \citep{ovadia2019,antoran2020depth}.
We note that, \citet{farquhar2020liberty} argue that in deeper networks the mean-field assumption should not be restrictive. 
Our empirical results seem to contradict this proposition. We find that scaling up approximations that \textit{do} consider weight correlations (e.g. \citet{mackay1992practical,louizos2016structured,maddox2019,ritter2018a}) by lowering the dimensionality of the weight space outperforms diagonal approximations. We conclude that more research is warranted in this area.
Finally, recent works have demonstrated the benefit of capturing the multi-modality of the posterior distribution via ensembles/mixtures \citep{lakshminarayanan2017,fort2019,filos2019benchmarking,wilson2020bayesian,eschenhagen2021mola}.

\textbf{Neural Linear Methods.}
These represent a generalised linear model in which the basis functions are defined by the $l{-}1$ first layers of a NN.
That is, neural linear methods perform inference over only the last layer of a NN, while keeping all other layers fixed \citep{snoek2015scalable,riquelme2018deep,ovadia2019,ober2019benchmarking,pinsler2019bayesian,kristiadi2020being}.
They can also be viewed as a special case of subnetwork inference, in which the subnetwork is simply defined to be the last NN layer.

\textbf{Inference over Subspaces.}
The subfield of NN pruning aims to increase the computational efficiency of NNs by identifying the smallest subset of weights which are required to make accurate predictions; see e.g.\ \cite{frankle2018the,Wang2020Picking}.
Our work differs in that it retains all NN weights but aims to find a small subset over which to perform probabilistic reasoning. %
More closely related work to ours is that of \cite{izmailov2019subspace}, who propose to perform inference over a low-dimensional subspace of weights; e.g.\ one constructed from the principal components of the SGD trajectory. Moreover, several recent approaches use low-rank parameterizations of approximate posteriors in the context of variational inference \citep{rossi2019walsh,swiatkowski2020k,dusenberry2020efficient}. This could also be viewed as doing inference over an implicit subspace of weight space.
In contrast, we propose a technique to find subsets of weights which are relevant to predictive uncertainty, i.e., we identify axis aligned subspaces.

\section{Conclusion}
Our work has three main findings: 1) modelling weight correlations in NNs is crucial to obtaining reliable predictive posteriors, 2) given these correlations, unimodal approximations of the posterior can be competitive with approximations that assign mass to multiple modes (e.g.\ deep ensembles), 3) inference does not need to be performed over all the weights in order to obtain reliable predictive posteriors.

We use these insights to develop a framework for scaling Bayesian inference to NNs with a large number of weights.
We approximate the posterior over a subset of the weights while keeping all others deterministic. 
Computational cost is decoupled from the total number of weights, allowing us to conveniently trade it off with the quality of approximation.
This allows us to use more expressive posterior approximations, such as full-covariance Gaussian distributions.

Linearized Laplace subnetwork inference can be applied post-hoc to any pre-trained model, making it particularly attractive for practical use.
Our empirical analysis suggests that this method 1) is more expressive and retains more uncertainty than crude approximations over the full network, 2) allows us to employ larger NNs, which fit a broader range of functions, without sacrificing the quality of our uncertainty estimates, and 3) is competitive with state-of-the-art uncertainty quantification methods, like deep ensembles.

We are excited to investigate combining subnetwork inference with different approximate inference methods, develop better subnetwork selection strategies and further explore the properties of subnetworks on the predictive distribution.

Finally, we are keen to apply recent advances to the Laplace approximation orthogonal to subnetwork inference, e.g.\ using the \emph{marginal likelihood} to select the prior precision \citep{immer2021scalable}, adding \emph{uncertainty units} to the model to improve its calibration \citep{kristiadi2021learnable}, or considering a \emph{mixture of Laplace approximations} to better capture the multi-modality of the posterior \citep{eschenhagen2021mola}.

\section*{Acknowledgments}
We thank Matthias Bauer, Alexander Immer, Andrew Y.\ K.\ Foong and Robert Pinsler for helpful discussions.
ED acknowledges funding from the EPSRC and Qualcomm.
JA acknowledges support from Microsoft Research, through its PhD Scholarship Programme, and from the EPSRC.
JUA acknowledges funding from the EPSRC and the Michael E. Fisher Studentship in Machine Learning.
This work has been performed using resources provided by the Cambridge Tier-2 system operated by the University of Cambridge Research Computing Service (http://www.hpc.cam.ac.uk) funded by EPSRC Tier-2 capital grant EP/P020259/1.

\bibliography{references}
\bibliographystyle{icml2021}

\onecolumn
\newpage
\appendix
\section{Additional Image Classification Results}
\label{sec:additional_results}

In this appendix, we provide additional experimental results for image classification tasks.

\subsection{Comparing the Parameter Efficiency of Subnetwork Linearized Laplace with Deep Ensembles}

Despite, the promising results shown by Subnetwork Linearized Laplace in \cref{sec:img_exp}, we note that our method has a notably larger space complexity than our baselines. We therefore investigate the parameter efficiency of our method. 

\begin{figure}
    \centering
    \includegraphics{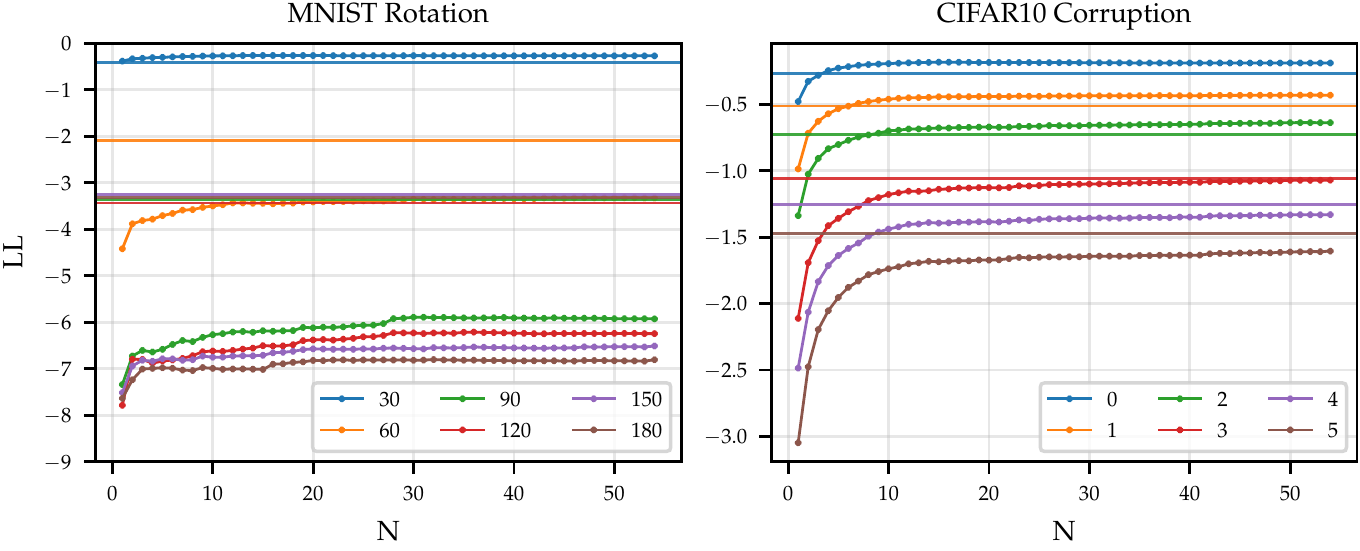}
    \vspace{-0.3cm}
    \caption{Rotated MNIST (left) and Corrupted CIFAR10 (right) results for deep ensembles \citep{lakshminarayanan2017} with large numbers of ensemble members (i.e. up to 55). Horizontal axis denotes number of ensemble members, and vertical axis denotes performance in terms of log-likelihood. Straight horizontal lines correspond to the performance of our method, as a reference. Colors denote different levels of rotation (left) and corruption (right).}
    \label{fig:ens_scale}
    \vspace{-0.3cm}
\end{figure}

Our ResNet18 Model has $\sim$11.2M parameters. Our subnetwork's covariance matrix contains 42,438$^{2}$ parameters. This totals $\sim$1,830M parameters. This same amount of memory could be used to store around 163 ensemble elements. In \cref{fig:ens_scale} we compare our subnetwork Linearized Laplace model with increasingly large ensembles on both rotated MNIST and corrupted CIFAR10. Although the performance of ensembles improves as more networks are added, it plateaus around 15 ensemble elements. This is in agreement with the findings of recent works \citep{antoran2020depth,ashukha2020pitfalls,lobacheva2020power}. At large rotations and corruptions, the log likelihood obtained by Subnetwork Linearised Laplace is greater than the asymptotic value obtained by ensembles. This suggests that using a larger number of parameters in a approximate posterior covariance matrix is a more efficient use of space than saving a large number of ensemble elements. We also note that inference in a very large ensemble requires performing a forward pass for every ensemble element. On the other hand, Linearised Laplace requires performing one backward pass for every output dimension and one forward pass.

\subsection{Scalability of Subnetwork Linearised Laplace in the number of Weights}

\begin{figure}
    \centering
    \includegraphics{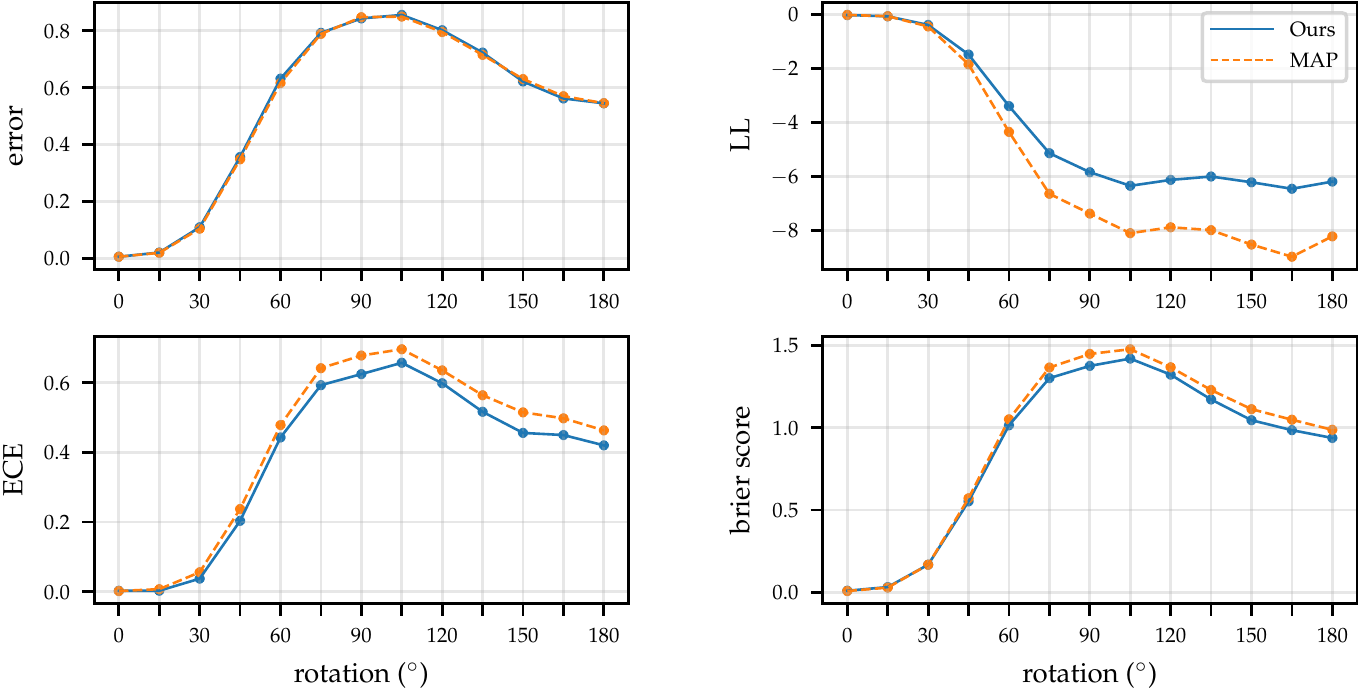}
    \vspace{-0.3cm}
    \caption{MNIST rotation results for ResNet-50, reporting predictive error, log-likelihood (LL), expected calibration error (ECE) and brier score. We choose a subnetwork containing only 0.167$\%$ (39,190 / 23,466,560) of the parameters of the full network. }
    \label{fig:mnist_res50}
    \vspace{-0.3cm}
\end{figure}

The aim of subnetwork inference is to scale existing posterior approximations to large networks. To further validate that this objective can be achieved, we perform subnetwork inference in ResNet50. We use a similar (slightly smaller) subnetwork size than we used with ResNet18: our subnetwork contains 39,190 / 23,466,560 ($0.167\%$) parameters. The results obtained with this model are displayed in \cref{fig:mnist_res50}. Subnetwork inference in ResNet50 improves upon a simple MAP estimate of the weights in terms of both log-likelihood and calibration metrics. %

\subsection{Out-of-Distribution Rejection}

In this section we provide additional results on out-of-distribution (OOD) rejection using predictive uncertainty. First, we train our models on a source dataset. We then evaluate them on the test set from our source dataset and on the test set of a target (out-of-distribution) dataset. We expect predictions for the target dataset to be more uncertain than those for the source dataset. Using predictive uncertainty as the discriminative variable we compute the area under ROC for each method under consideration and display them in \cref{tab:auc_roc}. The CIFAR-SVHN and MNIST-Fashion dataset pairs are chosen following \citet{nalisnick2018deep}. On the CIFAR-SVHN task, all methods perform similarly, except for ensembles, which clearly does best. On MNIST-Fashion, SWAG performs best, followed by Subnetwork Linearised Laplace and ensembles. 

\begin{table}[h]
    \centering
    \caption{AUC-ROC scores for out-of-distribution detection, using CIFAR10 vs SVHN and MNIST vs FashionMNIST as in- (source) and out-of-distribution (target) datasets, respectively.
    }
    \resizebox{\textwidth}{!}{\begin{tabular}{ll|lllllll}
    \toprule
      \textsc{Source} & \textsc{Target} &                 \textsc{Ours} &          \textsc{Ours (Rand)} &              \textsc{Dropout} &             \textsc{Diag-Lap} &             \textsc{Ensemble} &                  \textsc{MAP} &                 \textsc{SWAG} \\
    \midrule
    CIFAR10 & SVHN &  $0.85 \scriptstyle \pm 0.03$ &  $0.86 \scriptstyle \pm 0.02$ &  $0.85 \scriptstyle \pm 0.01$ &  $0.86 \scriptstyle \pm 0.02$ &  $0.91 \scriptstyle \pm 0.00$ &  $0.86 \scriptstyle \pm 0.02$ &  $0.83 \scriptstyle \pm 0.00$ \\
    MNIST & Fashion &  $0.92 \scriptstyle \pm 0.05$ &  $0.75 \scriptstyle \pm 0.02$ &  $0.82 \scriptstyle \pm 0.12$ &  $0.75 \scriptstyle \pm 0.01$ &  $0.90 \scriptstyle \pm 0.09$ &  $0.72 \scriptstyle \pm 0.03$ &  $0.97 \scriptstyle \pm 0.01$ \\
    \bottomrule
    \end{tabular}}
    \label{tab:auc_roc}
\end{table}

\begin{figure}[h]
    \centering
    \includegraphics{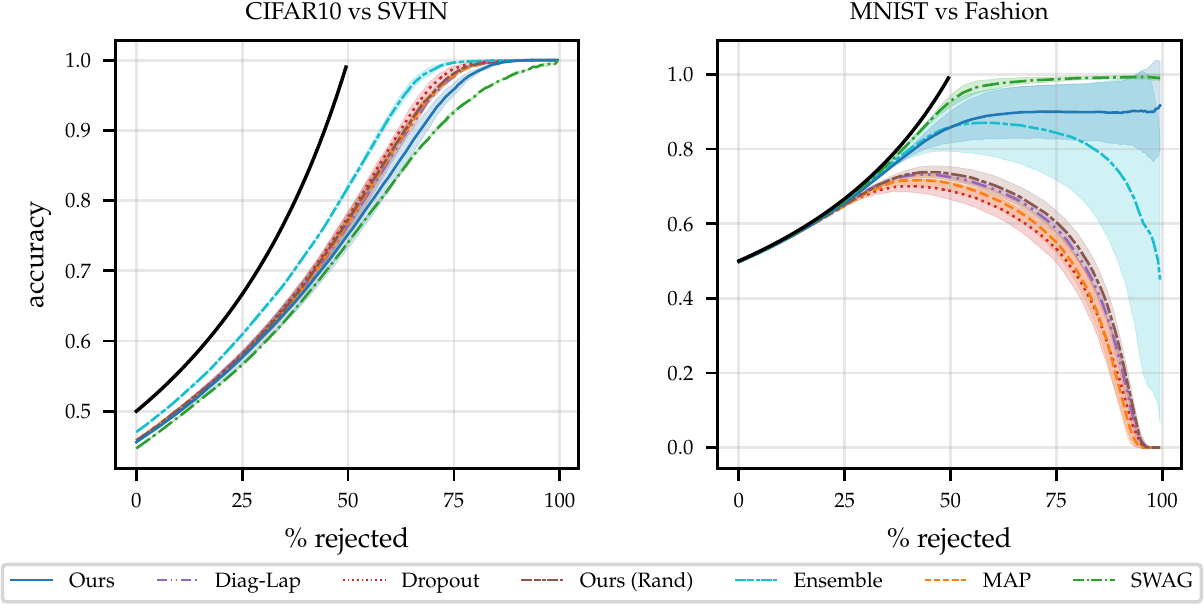}
    \vspace{-0.3cm}
    \caption{Rejection-classification plots.
    }
    \label{fig:rej_class}
    \vspace{-0.3cm}
\end{figure}

We also simulate a realistic OOD rejection scenario \citep{filos2019benchmarking} by jointly evaluating our models on an in-distribution and an OOD test set. We allow our methods to reject increasing proportions of the data based on predictive entropy before classifying the rest. All predictions on OOD samples are treated as incorrect. Following \citep{nalisnick2018deep}, we use CIFAR10 vs SVHN and MNIST vs FashionMNIST as in- and out-of-distribution datasets, respectively. Note that the SVHN test set is randomly sub-sampled down to a size of 10,000 to match that of CIFAR10. The results are shown in \cref{fig:rej_class}. On CIFAR-SVHN all methods perform similarly, with exceptions being ensembles, which perform best and SWAG which does worse. On MNIST-Fashion SWAG performs best, followed by Subnetwork Linearised Laplace. All other methods fail to distinguish very uncertain in-distribution data from low uncertainty OOD points. 

\clearpage

\subsection{Additional Rotation and Corruption Results}
\begin{figure}[H]
    \centering
    \includegraphics[width=0.76\textwidth]{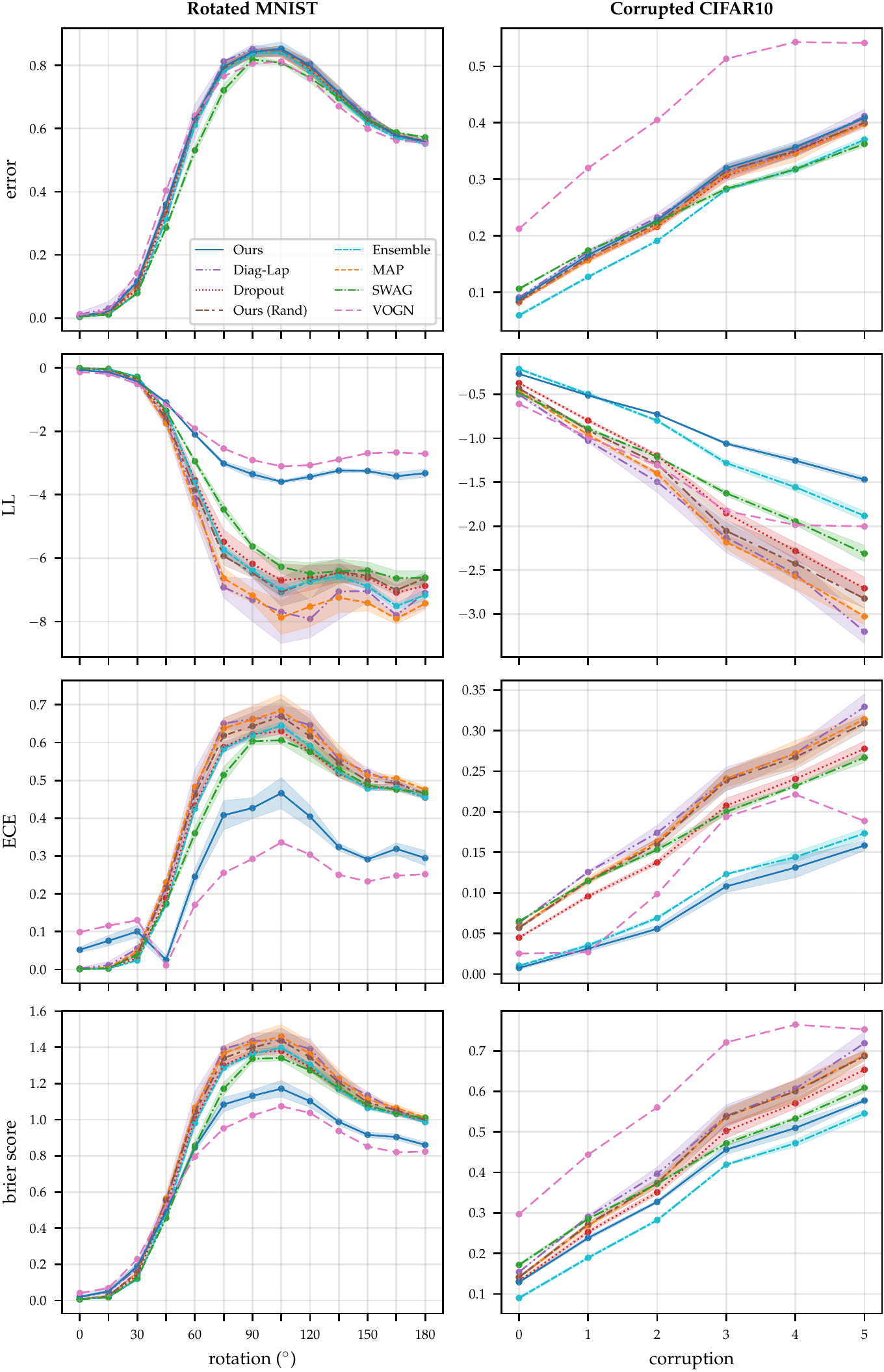}
    \vspace{-0.2cm}
    \caption{Full MNIST rotation and CIFAR10 corruption results, for ResNet-18, reporting predictive error, log-likelihood (LL), expected calibration error (ECE) and Brier score, respectively (from top to bottom).}
    \label{fig:img_res_full}
\end{figure}

We complement our results from \cref{fig:img_res} in the main text with results on additional calibration metrics: ECE and Brier Score, in \cref{fig:img_res_full}. Please refer to the appendix of \cite{antoran2020depth} for a description of these. 
\vspace{-0.3cm}

For reference, we provide our results from \cref{fig:img_res} and \cref{fig:img_res_full} in numerical format in the tables below. 

\begin{table}[h]
    \centering
    \caption{MNIST -- no rotation.}
    \resizebox{\textwidth}{!}{\begin{tabular}{l|rrrrrrrr}
    \toprule
     &                  \textsc{Ours} &           \textsc{Ours (Rand)} &               \textsc{Dropout} &              \textsc{Diag-Lap} &              \textsc{Ensemble} &                   \textsc{MAP} &                  \textsc{SWAG}  & \textsc{VOGN} \\
    \midrule
LL          &  $-0.07 \scriptstyle \pm 0.01$ &  $-0.01 \scriptstyle \pm 0.00$ &  $-0.01 \scriptstyle \pm 0.00$ &  $-0.04 \scriptstyle \pm 0.03$ &  $-0.01 \scriptstyle \pm 0.00$ &  $-0.01 \scriptstyle \pm 0.00$ &  $-0.01 \scriptstyle \pm 0.00$ &  $-0.14 \scriptstyle \pm nan$ \\
error       &   $0.01 \scriptstyle \pm 0.00$ &   $0.00 \scriptstyle \pm 0.00$ &   $0.00 \scriptstyle \pm 0.00$ &   $0.01 \scriptstyle \pm 0.01$ &   $0.00 \scriptstyle \pm 0.00$ &   $0.00 \scriptstyle \pm 0.00$ &   $0.00 \scriptstyle \pm 0.00$ &   $0.01 \scriptstyle \pm nan$ \\
ECE         &   $0.05 \scriptstyle \pm 0.01$ &   $0.00 \scriptstyle \pm 0.00$ &   $0.00 \scriptstyle \pm 0.00$ &   $0.00 \scriptstyle \pm 0.00$ &   $0.00 \scriptstyle \pm 0.00$ &   $0.00 \scriptstyle \pm 0.00$ &   $0.00 \scriptstyle \pm 0.00$ &   $0.10 \scriptstyle \pm nan$ \\
brier score &   $0.02 \scriptstyle \pm 0.00$ &   $0.01 \scriptstyle \pm 0.00$ &   $0.01 \scriptstyle \pm 0.00$ &   $0.02 \scriptstyle \pm 0.01$ &   $0.01 \scriptstyle \pm 0.00$ &   $0.01 \scriptstyle \pm 0.00$ &   $0.01 \scriptstyle \pm 0.00$ &   $0.04 \scriptstyle \pm nan$ \\
\bottomrule
    \end{tabular}}
    \label{tab:mnist_0}

    \caption{MNIST -- $15^{\circ}$ rotation.}
    \resizebox{\textwidth}{!}{\begin{tabular}{l|rrrrrrrr}
    \toprule
     &                  \textsc{Ours} &           \textsc{Ours (Rand)} &               \textsc{Dropout} &              \textsc{Diag-Lap} &              \textsc{Ensemble} &                   \textsc{MAP} &                  \textsc{SWAG}  & \textsc{VOGN} \\
    \midrule
LL          &  $-0.14 \scriptstyle \pm 0.02$ &  $-0.05 \scriptstyle \pm 0.00$ &  $-0.05 \scriptstyle \pm 0.00$ &  $-0.11 \scriptstyle \pm 0.08$ &  $-0.04 \scriptstyle \pm 0.00$ &  $-0.05 \scriptstyle \pm 0.00$ &  $-0.04 \scriptstyle \pm 0.00$ &  $-0.19 \scriptstyle \pm nan$ \\
error       &   $0.02 \scriptstyle \pm 0.00$ &   $0.02 \scriptstyle \pm 0.00$ &   $0.01 \scriptstyle \pm 0.00$ &   $0.03 \scriptstyle \pm 0.02$ &   $0.01 \scriptstyle \pm 0.00$ &   $0.02 \scriptstyle \pm 0.00$ &   $0.01 \scriptstyle \pm 0.00$ &   $0.02 \scriptstyle \pm nan$ \\
ECE         &   $0.08 \scriptstyle \pm 0.01$ &   $0.00 \scriptstyle \pm 0.00$ &   $0.00 \scriptstyle \pm 0.00$ &   $0.01 \scriptstyle \pm 0.01$ &   $0.00 \scriptstyle \pm 0.00$ &   $0.00 \scriptstyle \pm 0.00$ &   $0.00 \scriptstyle \pm 0.00$ &   $0.12 \scriptstyle \pm nan$ \\
brier score &   $0.05 \scriptstyle \pm 0.01$ &   $0.03 \scriptstyle \pm 0.00$ &   $0.02 \scriptstyle \pm 0.00$ &   $0.05 \scriptstyle \pm 0.03$ &   $0.02 \scriptstyle \pm 0.00$ &   $0.02 \scriptstyle \pm 0.00$ &   $0.02 \scriptstyle \pm 0.00$ &   $0.07 \scriptstyle \pm nan$ \\
\bottomrule
    \end{tabular}}
    \label{tab:mnist_15}

    \caption{MNIST -- $30^{\circ}$ rotation.}
    \resizebox{\textwidth}{!}{\begin{tabular}{l|rrrrrrrr}
    \toprule
     &                  \textsc{Ours} &           \textsc{Ours (Rand)} &               \textsc{Dropout} &              \textsc{Diag-Lap} &              \textsc{Ensemble} &                   \textsc{MAP} &                  \textsc{SWAG}  & \textsc{VOGN} \\
    \midrule
LL          &  $-0.42 \scriptstyle \pm 0.04$ &  $-0.36 \scriptstyle \pm 0.01$ &  $-0.32 \scriptstyle \pm 0.02$ &  $-0.44 \scriptstyle \pm 0.06$ &  $-0.28 \scriptstyle \pm 0.02$ &  $-0.39 \scriptstyle \pm 0.01$ &  $-0.30 \scriptstyle \pm 0.00$ &  $-0.51 \scriptstyle \pm nan$ \\
error       &   $0.11 \scriptstyle \pm 0.01$ &   $0.10 \scriptstyle \pm 0.00$ &   $0.09 \scriptstyle \pm 0.01$ &   $0.12 \scriptstyle \pm 0.01$ &   $0.08 \scriptstyle \pm 0.01$ &   $0.10 \scriptstyle \pm 0.00$ &   $0.08 \scriptstyle \pm 0.00$ &   $0.14 \scriptstyle \pm nan$ \\
ECE         &   $0.10 \scriptstyle \pm 0.02$ &   $0.04 \scriptstyle \pm 0.01$ &   $0.03 \scriptstyle \pm 0.00$ &   $0.06 \scriptstyle \pm 0.01$ &   $0.02 \scriptstyle \pm 0.00$ &   $0.05 \scriptstyle \pm 0.00$ &   $0.04 \scriptstyle \pm 0.00$ &   $0.13 \scriptstyle \pm nan$ \\
brier score &   $0.19 \scriptstyle \pm 0.02$ &   $0.16 \scriptstyle \pm 0.00$ &   $0.14 \scriptstyle \pm 0.01$ &   $0.18 \scriptstyle \pm 0.02$ &   $0.12 \scriptstyle \pm 0.01$ &   $0.16 \scriptstyle \pm 0.00$ &   $0.12 \scriptstyle \pm 0.00$ &   $0.23 \scriptstyle \pm nan$ \\
\bottomrule
    \end{tabular}}
    \label{tab:mnist_30}

    \caption{MNIST -- $45^{\circ}$ rotation.}
    \resizebox{\textwidth}{!}{\begin{tabular}{l|rrrrrrrr}
    \toprule
     &                  \textsc{Ours} &           \textsc{Ours (Rand)} &               \textsc{Dropout} &              \textsc{Diag-Lap} &              \textsc{Ensemble} &                   \textsc{MAP} &                  \textsc{SWAG}  & \textsc{VOGN} \\
    \midrule
LL          &  $-1.09 \scriptstyle \pm 0.03$ &  $-1.60 \scriptstyle \pm 0.05$ &  $-1.44 \scriptstyle \pm 0.11$ &  $-1.68 \scriptstyle \pm 0.20$ &  $-1.36 \scriptstyle \pm 0.07$ &  $-1.75 \scriptstyle \pm 0.06$ &  $-1.35 \scriptstyle \pm 0.02$ &  $-1.15 \scriptstyle \pm nan$ \\
error       &   $0.36 \scriptstyle \pm 0.01$ &   $0.35 \scriptstyle \pm 0.01$ &   $0.33 \scriptstyle \pm 0.01$ &   $0.35 \scriptstyle \pm 0.03$ &   $0.31 \scriptstyle \pm 0.01$ &   $0.35 \scriptstyle \pm 0.01$ &   $0.29 \scriptstyle \pm 0.00$ &   $0.40 \scriptstyle \pm nan$ \\
ECE         &   $0.03 \scriptstyle \pm 0.01$ &   $0.22 \scriptstyle \pm 0.01$ &   $0.19 \scriptstyle \pm 0.02$ &   $0.22 \scriptstyle \pm 0.02$ &   $0.17 \scriptstyle \pm 0.01$ &   $0.23 \scriptstyle \pm 0.01$ &   $0.18 \scriptstyle \pm 0.00$ &   $0.01 \scriptstyle \pm nan$ \\
brier score &   $0.49 \scriptstyle \pm 0.02$ &   $0.55 \scriptstyle \pm 0.02$ &   $0.52 \scriptstyle \pm 0.02$ &   $0.55 \scriptstyle \pm 0.04$ &   $0.48 \scriptstyle \pm 0.02$ &   $0.56 \scriptstyle \pm 0.02$ &   $0.46 \scriptstyle \pm 0.01$ &   $0.53 \scriptstyle \pm nan$ \\
\bottomrule
    \end{tabular}}
    \label{tab:mnist_45}

    \caption{MNIST -- $60^{\circ}$ rotation.}
    \resizebox{\textwidth}{!}{\begin{tabular}{l|rrrrrrrr}
    \toprule
     &                  \textsc{Ours} &           \textsc{Ours (Rand)} &               \textsc{Dropout} &              \textsc{Diag-Lap} &              \textsc{Ensemble} &                   \textsc{MAP} &                  \textsc{SWAG}  & \textsc{VOGN} \\
    \midrule
LL          &  $-2.10 \scriptstyle \pm 0.03$ &  $-3.85 \scriptstyle \pm 0.18$ &  $-3.54 \scriptstyle \pm 0.23$ &  $-4.11 \scriptstyle \pm 0.66$ &  $-3.60 \scriptstyle \pm 0.10$ &  $-4.29 \scriptstyle \pm 0.21$ &  $-2.95 \scriptstyle \pm 0.08$ &  $-1.92 \scriptstyle \pm nan$ \\
error       &   $0.63 \scriptstyle \pm 0.01$ &   $0.63 \scriptstyle \pm 0.01$ &   $0.62 \scriptstyle \pm 0.01$ &   $0.62 \scriptstyle \pm 0.05$ &   $0.61 \scriptstyle \pm 0.01$ &   $0.63 \scriptstyle \pm 0.01$ &   $0.53 \scriptstyle \pm 0.02$ &   $0.64 \scriptstyle \pm nan$ \\
ECE         &   $0.25 \scriptstyle \pm 0.02$ &   $0.46 \scriptstyle \pm 0.02$ &   $0.43 \scriptstyle \pm 0.02$ &   $0.47 \scriptstyle \pm 0.06$ &   $0.42 \scriptstyle \pm 0.01$ &   $0.48 \scriptstyle \pm 0.02$ &   $0.36 \scriptstyle \pm 0.02$ &   $0.17 \scriptstyle \pm nan$ \\
brier score &   $0.85 \scriptstyle \pm 0.02$ &   $1.04 \scriptstyle \pm 0.03$ &   $1.00 \scriptstyle \pm 0.03$ &   $1.05 \scriptstyle \pm 0.10$ &   $0.98 \scriptstyle \pm 0.02$ &   $1.07 \scriptstyle \pm 0.03$ &   $0.86 \scriptstyle \pm 0.03$ &   $0.80 \scriptstyle \pm nan$ \\
\bottomrule
    \end{tabular}}
    \label{tab:mnist_60}
\end{table}

\begin{table}[h]
    \centering
    \caption{MNIST -- $75^{\circ}$ rotation.}
    \resizebox{\textwidth}{!}{\begin{tabular}{l|rrrrrrrr}
    \toprule
     &                  \textsc{Ours} &           \textsc{Ours (Rand)} &               \textsc{Dropout} &              \textsc{Diag-Lap} &              \textsc{Ensemble} &                   \textsc{MAP} &                  \textsc{SWAG}  & \textsc{VOGN} \\
    \midrule
LL          &  $-3.02 \scriptstyle \pm 0.07$ &  $-5.93 \scriptstyle \pm 0.28$ &  $-5.49 \scriptstyle \pm 0.38$ &  $-6.92 \scriptstyle \pm 0.32$ &  $-5.74 \scriptstyle \pm 0.15$ &  $-6.63 \scriptstyle \pm 0.33$ &  $-4.46 \scriptstyle \pm 0.18$ &  $-2.54 \scriptstyle \pm nan$ \\
error       &   $0.80 \scriptstyle \pm 0.02$ &   $0.79 \scriptstyle \pm 0.01$ &   $0.79 \scriptstyle \pm 0.01$ &   $0.81 \scriptstyle \pm 0.00$ &   $0.78 \scriptstyle \pm 0.01$ &   $0.79 \scriptstyle \pm 0.01$ &   $0.72 \scriptstyle \pm 0.02$ &   $0.77 \scriptstyle \pm nan$ \\
ECE         &   $0.41 \scriptstyle \pm 0.04$ &   $0.62 \scriptstyle \pm 0.03$ &   $0.59 \scriptstyle \pm 0.01$ &   $0.65 \scriptstyle \pm 0.01$ &   $0.58 \scriptstyle \pm 0.01$ &   $0.64 \scriptstyle \pm 0.03$ &   $0.51 \scriptstyle \pm 0.02$ &   $0.26 \scriptstyle \pm nan$ \\
brier score &   $1.08 \scriptstyle \pm 0.04$ &   $1.34 \scriptstyle \pm 0.04$ &   $1.30 \scriptstyle \pm 0.02$ &   $1.39 \scriptstyle \pm 0.01$ &   $1.29 \scriptstyle \pm 0.02$ &   $1.37 \scriptstyle \pm 0.04$ &   $1.17 \scriptstyle \pm 0.04$ &   $0.95 \scriptstyle \pm nan$ \\
\bottomrule
    \end{tabular}}
    \label{tab:mnist_75}
\end{table}

\begin{table}[h]
    \centering
    \caption{MNIST -- $90^{\circ}$ rotation.}
    \resizebox{\textwidth}{!}{\begin{tabular}{l|rrrrrrrr}
    \toprule
     &                  \textsc{Ours} &           \textsc{Ours (Rand)} &               \textsc{Dropout} &              \textsc{Diag-Lap} &              \textsc{Ensemble} &                   \textsc{MAP} &                  \textsc{SWAG}  & \textsc{VOGN} \\
    \midrule
LL          &  $-3.35 \scriptstyle \pm 0.13$ &  $-6.46 \scriptstyle \pm 0.15$ &  $-6.18 \scriptstyle \pm 0.41$ &  $-7.32 \scriptstyle \pm 0.67$ &  $-6.39 \scriptstyle \pm 0.17$ &  $-7.18 \scriptstyle \pm 0.22$ &  $-5.63 \scriptstyle \pm 0.12$ &  $-2.91 \scriptstyle \pm nan$ \\
error       &   $0.84 \scriptstyle \pm 0.02$ &   $0.84 \scriptstyle \pm 0.01$ &   $0.84 \scriptstyle \pm 0.01$ &   $0.85 \scriptstyle \pm 0.01$ &   $0.84 \scriptstyle \pm 0.01$ &   $0.84 \scriptstyle \pm 0.01$ &   $0.82 \scriptstyle \pm 0.02$ &   $0.81 \scriptstyle \pm nan$ \\
ECE         &   $0.43 \scriptstyle \pm 0.03$ &   $0.64 \scriptstyle \pm 0.04$ &   $0.62 \scriptstyle \pm 0.01$ &   $0.66 \scriptstyle \pm 0.03$ &   $0.62 \scriptstyle \pm 0.01$ &   $0.66 \scriptstyle \pm 0.04$ &   $0.60 \scriptstyle \pm 0.01$ &   $0.29 \scriptstyle \pm nan$ \\
brier score &   $1.13 \scriptstyle \pm 0.03$ &   $1.40 \scriptstyle \pm 0.05$ &   $1.37 \scriptstyle \pm 0.01$ &   $1.44 \scriptstyle \pm 0.04$ &   $1.36 \scriptstyle \pm 0.01$ &   $1.43 \scriptstyle \pm 0.05$ &   $1.34 \scriptstyle \pm 0.02$ &   $1.02 \scriptstyle \pm nan$ \\
\bottomrule
    \end{tabular}}
    \label{tab:mnist_90}
\end{table}

\begin{table}[h]
    \centering
    \caption{MNIST -- $105^{\circ}$ rotation.}
    \resizebox{\textwidth}{!}{\begin{tabular}{l|rrrrrrrr}
    \toprule
     &                  \textsc{Ours} &           \textsc{Ours (Rand)} &               \textsc{Dropout} &              \textsc{Diag-Lap} &              \textsc{Ensemble} &                   \textsc{MAP} &                  \textsc{SWAG}  & \textsc{VOGN} \\
    \midrule
LL          &  $-3.59 \scriptstyle \pm 0.05$ &  $-7.06 \scriptstyle \pm 0.45$ &  $-6.70 \scriptstyle \pm 0.52$ &  $-7.69 \scriptstyle \pm 0.99$ &  $-7.01 \scriptstyle \pm 0.17$ &  $-7.87 \scriptstyle \pm 0.53$ &  $-6.28 \scriptstyle \pm 0.19$ &  $-3.10 \scriptstyle \pm nan$ \\
error       &   $0.85 \scriptstyle \pm 0.02$ &   $0.84 \scriptstyle \pm 0.02$ &   $0.84 \scriptstyle \pm 0.01$ &   $0.85 \scriptstyle \pm 0.01$ &   $0.84 \scriptstyle \pm 0.01$ &   $0.84 \scriptstyle \pm 0.02$ &   $0.81 \scriptstyle \pm 0.00$ &   $0.81 \scriptstyle \pm nan$ \\
ECE         &   $0.47 \scriptstyle \pm 0.04$ &   $0.67 \scriptstyle \pm 0.05$ &   $0.63 \scriptstyle \pm 0.01$ &   $0.67 \scriptstyle \pm 0.03$ &   $0.64 \scriptstyle \pm 0.01$ &   $0.68 \scriptstyle \pm 0.04$ &   $0.61 \scriptstyle \pm 0.01$ &   $0.34 \scriptstyle \pm nan$ \\
brier score &   $1.17 \scriptstyle \pm 0.05$ &   $1.44 \scriptstyle \pm 0.07$ &   $1.38 \scriptstyle \pm 0.02$ &   $1.44 \scriptstyle \pm 0.04$ &   $1.40 \scriptstyle \pm 0.01$ &   $1.46 \scriptstyle \pm 0.07$ &   $1.34 \scriptstyle \pm 0.02$ &   $1.07 \scriptstyle \pm nan$ \\
\bottomrule
    \end{tabular}}
    \label{tab:mnist_105}
\end{table}

\begin{table}[h]
    \centering
    \caption{MNIST -- $120^{\circ}$ rotation.}
    \resizebox{\textwidth}{!}{\begin{tabular}{l|rrrrrrrr}
    \toprule
     &                  \textsc{Ours} &           \textsc{Ours (Rand)} &               \textsc{Dropout} &              \textsc{Diag-Lap} &              \textsc{Ensemble} &                   \textsc{MAP} &                  \textsc{SWAG}  & \textsc{VOGN} \\
    \midrule
LL          &  $-3.43 \scriptstyle \pm 0.07$ &  $-6.73 \scriptstyle \pm 0.53$ &  $-6.62 \scriptstyle \pm 0.39$ &  $-7.92 \scriptstyle \pm 0.59$ &  $-6.73 \scriptstyle \pm 0.11$ &  $-7.53 \scriptstyle \pm 0.63$ &  $-6.49 \scriptstyle \pm 0.36$ &  $-3.07 \scriptstyle \pm nan$ \\
error       &   $0.80 \scriptstyle \pm 0.02$ &   $0.79 \scriptstyle \pm 0.02$ &   $0.78 \scriptstyle \pm 0.01$ &   $0.81 \scriptstyle \pm 0.01$ &   $0.78 \scriptstyle \pm 0.01$ &   $0.79 \scriptstyle \pm 0.02$ &   $0.76 \scriptstyle \pm 0.02$ &   $0.76 \scriptstyle \pm nan$ \\
ECE         &   $0.40 \scriptstyle \pm 0.03$ &   $0.62 \scriptstyle \pm 0.05$ &   $0.58 \scriptstyle \pm 0.01$ &   $0.65 \scriptstyle \pm 0.04$ &   $0.59 \scriptstyle \pm 0.01$ &   $0.63 \scriptstyle \pm 0.04$ &   $0.58 \scriptstyle \pm 0.03$ &   $0.30 \scriptstyle \pm nan$ \\
brier score &   $1.10 \scriptstyle \pm 0.03$ &   $1.35 \scriptstyle \pm 0.07$ &   $1.29 \scriptstyle \pm 0.02$ &   $1.39 \scriptstyle \pm 0.06$ &   $1.30 \scriptstyle \pm 0.01$ &   $1.36 \scriptstyle \pm 0.07$ &   $1.27 \scriptstyle \pm 0.04$ &   $1.04 \scriptstyle \pm nan$ \\
\bottomrule
    \end{tabular}}
    \label{tab:mnist_120}
\end{table}

\begin{table}[h]
    \centering
    \caption{MNIST -- $135^{\circ}$ rotation.}
    \resizebox{\textwidth}{!}{\begin{tabular}{l|rrrrrrrr}
    \toprule
     &                  \textsc{Ours} &           \textsc{Ours (Rand)} &               \textsc{Dropout} &              \textsc{Diag-Lap} &              \textsc{Ensemble} &                   \textsc{MAP} &                  \textsc{SWAG}  & \textsc{VOGN} \\
    \midrule
LL          &  $-3.24 \scriptstyle \pm 0.06$ &  $-6.43 \scriptstyle \pm 0.38$ &  $-6.46 \scriptstyle \pm 0.28$ &  $-7.05 \scriptstyle \pm 0.88$ &  $-6.57 \scriptstyle \pm 0.10$ &  $-7.24 \scriptstyle \pm 0.48$ &  $-6.40 \scriptstyle \pm 0.37$ &  $-2.89 \scriptstyle \pm nan$ \\
error       &   $0.71 \scriptstyle \pm 0.02$ &   $0.71 \scriptstyle \pm 0.02$ &   $0.70 \scriptstyle \pm 0.01$ &   $0.71 \scriptstyle \pm 0.01$ &   $0.70 \scriptstyle \pm 0.01$ &   $0.71 \scriptstyle \pm 0.02$ &   $0.70 \scriptstyle \pm 0.02$ &   $0.67 \scriptstyle \pm nan$ \\
ECE         &   $0.32 \scriptstyle \pm 0.01$ &   $0.55 \scriptstyle \pm 0.03$ &   $0.52 \scriptstyle \pm 0.01$ &   $0.56 \scriptstyle \pm 0.02$ &   $0.52 \scriptstyle \pm 0.01$ &   $0.56 \scriptstyle \pm 0.03$ &   $0.53 \scriptstyle \pm 0.02$ &   $0.25 \scriptstyle \pm nan$ \\
brier score &   $0.99 \scriptstyle \pm 0.02$ &   $1.21 \scriptstyle \pm 0.05$ &   $1.17 \scriptstyle \pm 0.02$ &   $1.22 \scriptstyle \pm 0.04$ &   $1.17 \scriptstyle \pm 0.01$ &   $1.23 \scriptstyle \pm 0.05$ &   $1.18 \scriptstyle \pm 0.04$ &   $0.94 \scriptstyle \pm nan$ \\
\bottomrule
    \end{tabular}}
    \label{tab:mnist_135}
\end{table}

\begin{table}[h]
    \centering
    \caption{MNIST -- $150^{\circ}$ rotation.}
    \resizebox{\textwidth}{!}{\begin{tabular}{l|rrrrrrrr}
    \toprule
     &                  \textsc{Ours} &           \textsc{Ours (Rand)} &               \textsc{Dropout} &              \textsc{Diag-Lap} &              \textsc{Ensemble} &                   \textsc{MAP} &                  \textsc{SWAG}  & \textsc{VOGN} \\
    \midrule
LL          &  $-3.25 \scriptstyle \pm 0.05$ &  $-6.56 \scriptstyle \pm 0.18$ &  $-6.62 \scriptstyle \pm 0.33$ &  $-7.04 \scriptstyle \pm 0.36$ &  $-6.88 \scriptstyle \pm 0.11$ &  $-7.41 \scriptstyle \pm 0.25$ &  $-6.39 \scriptstyle \pm 0.27$ &  $-2.69 \scriptstyle \pm nan$ \\
error       &   $0.63 \scriptstyle \pm 0.02$ &   $0.63 \scriptstyle \pm 0.01$ &   $0.63 \scriptstyle \pm 0.00$ &   $0.65 \scriptstyle \pm 0.01$ &   $0.62 \scriptstyle \pm 0.01$ &   $0.63 \scriptstyle \pm 0.01$ &   $0.63 \scriptstyle \pm 0.01$ &   $0.60 \scriptstyle \pm nan$ \\
ECE         &   $0.29 \scriptstyle \pm 0.01$ &   $0.50 \scriptstyle \pm 0.01$ &   $0.48 \scriptstyle \pm 0.01$ &   $0.52 \scriptstyle \pm 0.01$ &   $0.48 \scriptstyle \pm 0.01$ &   $0.51 \scriptstyle \pm 0.01$ &   $0.49 \scriptstyle \pm 0.01$ &   $0.23 \scriptstyle \pm nan$ \\
brier score &   $0.92 \scriptstyle \pm 0.02$ &   $1.10 \scriptstyle \pm 0.02$ &   $1.07 \scriptstyle \pm 0.01$ &   $1.13 \scriptstyle \pm 0.02$ &   $1.06 \scriptstyle \pm 0.01$ &   $1.11 \scriptstyle \pm 0.02$ &   $1.08 \scriptstyle \pm 0.02$ &   $0.85 \scriptstyle \pm nan$ \\
\bottomrule
    \end{tabular}}
    \label{tab:mnist_150}
\end{table}

\begin{table}[h]
    \centering
    \caption{MNIST -- $165^{\circ}$ rotation.}
    \resizebox{\textwidth}{!}{\begin{tabular}{l|rrrrrrrr}
    \toprule
     &                  \textsc{Ours} &           \textsc{Ours (Rand)} &               \textsc{Dropout} &              \textsc{Diag-Lap} &              \textsc{Ensemble} &                   \textsc{MAP} &                  \textsc{SWAG}  & \textsc{VOGN} \\
    \midrule
LL          &  $-3.42 \scriptstyle \pm 0.12$ &  $-7.01 \scriptstyle \pm 0.15$ &  $-7.08 \scriptstyle \pm 0.39$ &  $-7.80 \scriptstyle \pm 0.12$ &  $-7.51 \scriptstyle \pm 0.11$ &  $-7.91 \scriptstyle \pm 0.18$ &  $-6.63 \scriptstyle \pm 0.24$ &  $-2.67 \scriptstyle \pm nan$ \\
error       &   $0.58 \scriptstyle \pm 0.01$ &   $0.58 \scriptstyle \pm 0.01$ &   $0.58 \scriptstyle \pm 0.01$ &   $0.58 \scriptstyle \pm 0.00$ &   $0.57 \scriptstyle \pm 0.01$ &   $0.58 \scriptstyle \pm 0.01$ &   $0.59 \scriptstyle \pm 0.00$ &   $0.56 \scriptstyle \pm nan$ \\
ECE         &   $0.32 \scriptstyle \pm 0.02$ &   $0.49 \scriptstyle \pm 0.01$ &   $0.48 \scriptstyle \pm 0.01$ &   $0.49 \scriptstyle \pm 0.01$ &   $0.48 \scriptstyle \pm 0.00$ &   $0.51 \scriptstyle \pm 0.01$ &   $0.48 \scriptstyle \pm 0.00$ &   $0.25 \scriptstyle \pm nan$ \\
brier score &   $0.90 \scriptstyle \pm 0.02$ &   $1.05 \scriptstyle \pm 0.01$ &   $1.04 \scriptstyle \pm 0.01$ &   $1.05 \scriptstyle \pm 0.01$ &   $1.03 \scriptstyle \pm 0.01$ &   $1.07 \scriptstyle \pm 0.02$ &   $1.03 \scriptstyle \pm 0.01$ &   $0.82 \scriptstyle \pm nan$ \\
\bottomrule
    \end{tabular}}
    \label{tab:mnist_165}
\end{table}

\begin{table}[h]
    \centering
    \caption{MNIST -- $180^{\circ}$ rotation.}
    \resizebox{\textwidth}{!}{\begin{tabular}{l|rrrrrrrr}
    \toprule
     &                  \textsc{Ours} &           \textsc{Ours (Rand)} &               \textsc{Dropout} &              \textsc{Diag-Lap} &              \textsc{Ensemble} &                   \textsc{MAP} &                  \textsc{SWAG}  & \textsc{VOGN} \\
    \midrule
LL          &  $-3.32 \scriptstyle \pm 0.13$ &  $-6.63 \scriptstyle \pm 0.18$ &  $-6.87 \scriptstyle \pm 0.32$ &  $-7.10 \scriptstyle \pm 0.47$ &  $-7.16 \scriptstyle \pm 0.16$ &  $-7.43 \scriptstyle \pm 0.20$ &  $-6.61 \scriptstyle \pm 0.22$ &  $-2.71 \scriptstyle \pm nan$ \\
error       &   $0.56 \scriptstyle \pm 0.01$ &   $0.56 \scriptstyle \pm 0.01$ &   $0.56 \scriptstyle \pm 0.00$ &   $0.55 \scriptstyle \pm 0.01$ &   $0.55 \scriptstyle \pm 0.00$ &   $0.56 \scriptstyle \pm 0.01$ &   $0.57 \scriptstyle \pm 0.00$ &   $0.55 \scriptstyle \pm nan$ \\
ECE         &   $0.29 \scriptstyle \pm 0.02$ &   $0.46 \scriptstyle \pm 0.01$ &   $0.45 \scriptstyle \pm 0.00$ &   $0.46 \scriptstyle \pm 0.00$ &   $0.46 \scriptstyle \pm 0.01$ &   $0.48 \scriptstyle \pm 0.01$ &   $0.47 \scriptstyle \pm 0.01$ &   $0.25 \scriptstyle \pm nan$ \\
brier score &   $0.86 \scriptstyle \pm 0.02$ &   $1.00 \scriptstyle \pm 0.01$ &   $0.99 \scriptstyle \pm 0.01$ &   $0.99 \scriptstyle \pm 0.01$ &   $0.99 \scriptstyle \pm 0.00$ &   $1.01 \scriptstyle \pm 0.02$ &   $1.01 \scriptstyle \pm 0.01$ &   $0.82 \scriptstyle \pm nan$ \\
\bottomrule
    \end{tabular}}
    \label{tab:mnist_180}
\end{table}

\begin{table}[h]
    \centering
    \caption{CIFAR10 -- no corruption.}
    \resizebox{\textwidth}{!}{\begin{tabular}{l|rrrrrrrr}
    \toprule
     &                  \textsc{Ours} &           \textsc{Ours (Rand)} &               \textsc{Dropout} &              \textsc{Diag-Lap} &              \textsc{Ensemble} &                   \textsc{MAP} &                  \textsc{SWAG}  & \textsc{VOGN} \\
    \midrule
LL          &  $-0.27 \scriptstyle \pm 0.00$ &  $-0.43 \scriptstyle \pm 0.01$ &  $-0.37 \scriptstyle \pm 0.01$ &  $-0.50 \scriptstyle \pm 0.02$ &  $-0.21 \scriptstyle \pm 0.01$ &  $-0.46 \scriptstyle \pm 0.02$ &  $-0.48 \scriptstyle \pm 0.01$ &  $-0.61 \scriptstyle \pm nan$ \\
error       &   $0.09 \scriptstyle \pm 0.00$ &   $0.08 \scriptstyle \pm 0.00$ &   $0.08 \scriptstyle \pm 0.00$ &   $0.09 \scriptstyle \pm 0.00$ &   $0.06 \scriptstyle \pm 0.00$ &   $0.08 \scriptstyle \pm 0.00$ &   $0.11 \scriptstyle \pm 0.00$ &   $0.21 \scriptstyle \pm nan$ \\
ECE         &   $0.01 \scriptstyle \pm 0.00$ &   $0.06 \scriptstyle \pm 0.00$ &   $0.04 \scriptstyle \pm 0.00$ &   $0.06 \scriptstyle \pm 0.00$ &   $0.01 \scriptstyle \pm 0.00$ &   $0.06 \scriptstyle \pm 0.00$ &   $0.07 \scriptstyle \pm 0.00$ &   $0.03 \scriptstyle \pm nan$ \\
brier score &   $0.13 \scriptstyle \pm 0.00$ &   $0.14 \scriptstyle \pm 0.00$ &   $0.13 \scriptstyle \pm 0.00$ &   $0.15 \scriptstyle \pm 0.00$ &   $0.09 \scriptstyle \pm 0.00$ &   $0.14 \scriptstyle \pm 0.00$ &   $0.17 \scriptstyle \pm 0.00$ &   $0.30 \scriptstyle \pm nan$ \\
\bottomrule
    \end{tabular}}
    \label{tab:cifar_0}
\end{table}

\begin{table}[h]
    \centering
    \caption{CIFAR10 -- level $1$ corruption.}
    \resizebox{\textwidth}{!}{\begin{tabular}{l|rrrrrrrr}
    \toprule
     &                  \textsc{Ours} &           \textsc{Ours (Rand)} &               \textsc{Dropout} &              \textsc{Diag-Lap} &              \textsc{Ensemble} &                   \textsc{MAP} &                  \textsc{SWAG}  & \textsc{VOGN} \\
    \midrule
LL          &  $-0.51 \scriptstyle \pm 0.01$ &  $-0.91 \scriptstyle \pm 0.01$ &  $-0.80 \scriptstyle \pm 0.02$ &  $-1.03 \scriptstyle \pm 0.02$ &  $-0.50 \scriptstyle \pm 0.02$ &  $-0.96 \scriptstyle \pm 0.02$ &  $-0.89 \scriptstyle \pm 0.02$ &  $-0.99 \scriptstyle \pm nan$ \\
error       &   $0.17 \scriptstyle \pm 0.01$ &   $0.16 \scriptstyle \pm 0.00$ &   $0.16 \scriptstyle \pm 0.00$ &   $0.17 \scriptstyle \pm 0.00$ &   $0.13 \scriptstyle \pm 0.00$ &   $0.16 \scriptstyle \pm 0.00$ &   $0.17 \scriptstyle \pm 0.00$ &   $0.32 \scriptstyle \pm nan$ \\
ECE         &   $0.03 \scriptstyle \pm 0.00$ &   $0.11 \scriptstyle \pm 0.00$ &   $0.10 \scriptstyle \pm 0.00$ &   $0.13 \scriptstyle \pm 0.00$ &   $0.04 \scriptstyle \pm 0.00$ &   $0.12 \scriptstyle \pm 0.01$ &   $0.11 \scriptstyle \pm 0.00$ &   $0.03 \scriptstyle \pm nan$ \\
brier score &   $0.24 \scriptstyle \pm 0.00$ &   $0.27 \scriptstyle \pm 0.00$ &   $0.25 \scriptstyle \pm 0.00$ &   $0.29 \scriptstyle \pm 0.00$ &   $0.19 \scriptstyle \pm 0.00$ &   $0.27 \scriptstyle \pm 0.01$ &   $0.29 \scriptstyle \pm 0.00$ &   $0.44 \scriptstyle \pm nan$ \\
\bottomrule
    \end{tabular}}
    \label{tab:cifar_1}
\end{table}

\begin{table}[h]
    \centering
    \caption{CIFAR10 -- level $2$ corruption.}
    \resizebox{\textwidth}{!}{\begin{tabular}{l|rrrrrrrr}
    \toprule
     &                  \textsc{Ours} &           \textsc{Ours (Rand)} &               \textsc{Dropout} &              \textsc{Diag-Lap} &              \textsc{Ensemble} &                   \textsc{MAP} &                  \textsc{SWAG}  & \textsc{VOGN} \\
    \midrule
LL          &  $-0.73 \scriptstyle \pm 0.01$ &  $-1.29 \scriptstyle \pm 0.06$ &  $-1.20 \scriptstyle \pm 0.02$ &  $-1.50 \scriptstyle \pm 0.12$ &  $-0.80 \scriptstyle \pm 0.01$ &  $-1.40 \scriptstyle \pm 0.03$ &  $-1.21 \scriptstyle \pm 0.00$ &  $-1.31 \scriptstyle \pm nan$ \\
error       &   $0.23 \scriptstyle \pm 0.00$ &   $0.22 \scriptstyle \pm 0.01$ &   $0.22 \scriptstyle \pm 0.00$ &   $0.23 \scriptstyle \pm 0.01$ &   $0.19 \scriptstyle \pm 0.00$ &   $0.22 \scriptstyle \pm 0.00$ &   $0.22 \scriptstyle \pm 0.00$ &   $0.40 \scriptstyle \pm nan$ \\
ECE         &   $0.06 \scriptstyle \pm 0.00$ &   $0.16 \scriptstyle \pm 0.01$ &   $0.14 \scriptstyle \pm 0.00$ &   $0.17 \scriptstyle \pm 0.01$ &   $0.07 \scriptstyle \pm 0.00$ &   $0.16 \scriptstyle \pm 0.00$ &   $0.15 \scriptstyle \pm 0.00$ &   $0.10 \scriptstyle \pm nan$ \\
brier score &   $0.33 \scriptstyle \pm 0.00$ &   $0.37 \scriptstyle \pm 0.01$ &   $0.35 \scriptstyle \pm 0.01$ &   $0.40 \scriptstyle \pm 0.02$ &   $0.28 \scriptstyle \pm 0.00$ &   $0.37 \scriptstyle \pm 0.01$ &   $0.37 \scriptstyle \pm 0.00$ &   $0.56 \scriptstyle \pm nan$ \\
\bottomrule
    \end{tabular}}
    \label{tab:cifar_2}
\end{table}

\begin{table}[h]
    \centering
    \caption{CIFAR10 -- level $3$ corruption.}
    \resizebox{\textwidth}{!}{\begin{tabular}{l|rrrrrrrr}
    \toprule
     &                  \textsc{Ours} &           \textsc{Ours (Rand)} &               \textsc{Dropout} &              \textsc{Diag-Lap} &              \textsc{Ensemble} &                   \textsc{MAP} &                  \textsc{SWAG}  & \textsc{VOGN} \\
    \midrule
LL          &  $-1.06 \scriptstyle \pm 0.02$ &  $-2.06 \scriptstyle \pm 0.12$ &  $-1.85 \scriptstyle \pm 0.07$ &  $-2.13 \scriptstyle \pm 0.17$ &  $-1.28 \scriptstyle \pm 0.03$ &  $-2.18 \scriptstyle \pm 0.08$ &  $-1.63 \scriptstyle \pm 0.03$ &  $-1.83 \scriptstyle \pm nan$ \\
error       &   $0.32 \scriptstyle \pm 0.01$ &   $0.31 \scriptstyle \pm 0.01$ &   $0.31 \scriptstyle \pm 0.01$ &   $0.31 \scriptstyle \pm 0.01$ &   $0.28 \scriptstyle \pm 0.00$ &   $0.31 \scriptstyle \pm 0.01$ &   $0.28 \scriptstyle \pm 0.00$ &   $0.51 \scriptstyle \pm nan$ \\
ECE         &   $0.11 \scriptstyle \pm 0.01$ &   $0.24 \scriptstyle \pm 0.01$ &   $0.21 \scriptstyle \pm 0.01$ &   $0.24 \scriptstyle \pm 0.01$ &   $0.12 \scriptstyle \pm 0.00$ &   $0.24 \scriptstyle \pm 0.01$ &   $0.20 \scriptstyle \pm 0.00$ &   $0.19 \scriptstyle \pm nan$ \\
brier score &   $0.46 \scriptstyle \pm 0.01$ &   $0.54 \scriptstyle \pm 0.02$ &   $0.50 \scriptstyle \pm 0.02$ &   $0.54 \scriptstyle \pm 0.03$ &   $0.42 \scriptstyle \pm 0.00$ &   $0.54 \scriptstyle \pm 0.02$ &   $0.47 \scriptstyle \pm 0.01$ &   $0.72 \scriptstyle \pm nan$ \\
\bottomrule
    \end{tabular}}
    \label{tab:cifar_3}
\end{table}

\begin{table}[h]
    \centering
    \caption{CIFAR10 -- level $4$ corruption.}
    \resizebox{\textwidth}{!}{\begin{tabular}{l|rrrrrrrr}
    \toprule
     &                  \textsc{Ours} &           \textsc{Ours (Rand)} &               \textsc{Dropout} &              \textsc{Diag-Lap} &              \textsc{Ensemble} &                   \textsc{MAP} &                  \textsc{SWAG}  & \textsc{VOGN} \\
    \midrule
LL          &  $-1.25 \scriptstyle \pm 0.03$ &  $-2.43 \scriptstyle \pm 0.18$ &  $-2.28 \scriptstyle \pm 0.10$ &  $-2.54 \scriptstyle \pm 0.18$ &  $-1.56 \scriptstyle \pm 0.05$ &  $-2.57 \scriptstyle \pm 0.15$ &  $-1.95 \scriptstyle \pm 0.04$ &  $-1.99 \scriptstyle \pm nan$ \\
error       &   $0.36 \scriptstyle \pm 0.01$ &   $0.35 \scriptstyle \pm 0.01$ &   $0.35 \scriptstyle \pm 0.01$ &   $0.35 \scriptstyle \pm 0.01$ &   $0.32 \scriptstyle \pm 0.01$ &   $0.35 \scriptstyle \pm 0.01$ &   $0.32 \scriptstyle \pm 0.00$ &   $0.54 \scriptstyle \pm nan$ \\
ECE         &   $0.13 \scriptstyle \pm 0.01$ &   $0.27 \scriptstyle \pm 0.01$ &   $0.24 \scriptstyle \pm 0.01$ &   $0.27 \scriptstyle \pm 0.01$ &   $0.14 \scriptstyle \pm 0.01$ &   $0.27 \scriptstyle \pm 0.02$ &   $0.23 \scriptstyle \pm 0.00$ &   $0.22 \scriptstyle \pm nan$ \\
brier score &   $0.51 \scriptstyle \pm 0.02$ &   $0.60 \scriptstyle \pm 0.03$ &   $0.57 \scriptstyle \pm 0.01$ &   $0.61 \scriptstyle \pm 0.02$ &   $0.47 \scriptstyle \pm 0.01$ &   $0.60 \scriptstyle \pm 0.03$ &   $0.53 \scriptstyle \pm 0.00$ &   $0.76 \scriptstyle \pm nan$ \\
\bottomrule
    \end{tabular}}
    \label{tab:cifar_4}
\end{table}

\begin{table}[h]
    \centering
    \caption{CIFAR10 -- level $5$ corruption.}
    \resizebox{\textwidth}{!}{\begin{tabular}{l|rrrrrrrr}
    \toprule
     &                  \textsc{Ours} &           \textsc{Ours (Rand)} &               \textsc{Dropout} &              \textsc{Diag-Lap} &              \textsc{Ensemble} &                   \textsc{MAP} &                  \textsc{SWAG}  & \textsc{VOGN} \\
    \midrule
LL          &  $-1.47 \scriptstyle \pm 0.03$ &  $-2.82 \scriptstyle \pm 0.11$ &  $-2.71 \scriptstyle \pm 0.13$ &  $-3.20 \scriptstyle \pm 0.13$ &  $-1.88 \scriptstyle \pm 0.05$ &  $-3.03 \scriptstyle \pm 0.10$ &  $-2.31 \scriptstyle \pm 0.09$ &  $-2.00 \scriptstyle \pm nan$ \\
error       &   $0.41 \scriptstyle \pm 0.00$ &   $0.40 \scriptstyle \pm 0.01$ &   $0.40 \scriptstyle \pm 0.01$ &   $0.41 \scriptstyle \pm 0.01$ &   $0.37 \scriptstyle \pm 0.01$ &   $0.40 \scriptstyle \pm 0.00$ &   $0.36 \scriptstyle \pm 0.01$ &   $0.54 \scriptstyle \pm nan$ \\
ECE         &   $0.16 \scriptstyle \pm 0.01$ &   $0.31 \scriptstyle \pm 0.01$ &   $0.28 \scriptstyle \pm 0.01$ &   $0.33 \scriptstyle \pm 0.02$ &   $0.17 \scriptstyle \pm 0.01$ &   $0.31 \scriptstyle \pm 0.01$ &   $0.27 \scriptstyle \pm 0.01$ &   $0.19 \scriptstyle \pm nan$ \\
brier score &   $0.58 \scriptstyle \pm 0.00$ &   $0.69 \scriptstyle \pm 0.01$ &   $0.65 \scriptstyle \pm 0.01$ &   $0.72 \scriptstyle \pm 0.03$ &   $0.55 \scriptstyle \pm 0.01$ &   $0.69 \scriptstyle \pm 0.01$ &   $0.61 \scriptstyle \pm 0.01$ &   $0.75 \scriptstyle \pm nan$ \\
\bottomrule
    \end{tabular}}
    \label{tab:cifar_5}
\end{table}

\FloatBarrier
\section{Derivations for the Wasserstein Pruning Objective}
\label{sec:proofs}

\subsection{Derivation for \cref{eq:wass2squared}}
Note that, for our linearized model (described in \cref{sec:lin_laplace}), the true posterior $p(\rvw | \mathcal{D})$ is either Gaussian or approximately Gaussian. Additionally, the approximate posterior $q_S(\rvw) = q(\rvw_S) \ \prod_{r} \delta(\rw_{r} - \widehat{\rw}_{r} )$ can be seen as a degenerate Gaussian in which rows and columns of the covariance matrix are zeroed out.
Thus, we consider the squared 2-Wasserstein distance between two Gaussian distributions $\mathcal{N}(\vmu_1, \mSigma_1)$ and $\mathcal{N}(\vmu_2, \mSigma_2)$, which has the following closed-form expression \citep{givens1984class}\footnote{This also holds for our case of a degenerate Gaussian with singular covariance matrix \citep{givens1984class}.}:
\begin{equation}
    W_2\left( \mathcal{N}\left(\vmu_1, \mSigma_1\right),  \mathcal{N}\left(\vmu_2, \mSigma_2\right) \right)^2 = \|\vmu_{1} - \vmu_{2}\|_2^2 + \text{Tr}\left(\mSigma_1 + \mSigma_2 - 2\left(\mSigma_2^{1/2} \mSigma_1 \mSigma_2^{1/2} \right)^{1/2}\right)\ .
\end{equation}
In this case both distributions have the same mean: $\vmu_1 = \vmu_2 = \widehat{\rvw}$. The true posterior's covariance matrix is the inverse GGN matrix, i.e. $\mSigma_1 = \widetilde{\rmH}^{-1}$. For the approximate posterior $\mSigma_2 = \widetilde{\rmH}_{S+}^{-1}$, which is equal to $\widetilde{\rmH}_{S}^{-1}$ (the inverse GGN matrix of the subnetwork) padded with zeros at the positions corresponding to point estimated weights $\rw_r$, matching the shape of $\widetilde{\rmH}^{-1}$. Alternatively, but equivalently, we can define $\widetilde{\rmH}_{S+}^{-1} = \rmM_S \odot \widetilde{\rmH}^{-1}$, where $\odot$ is the Hadamard product, and $\rmM_S$ is a mask matrix with zeros in the rows and columns corresponding to $\rw_r$, i.e. the rows and columns corresponding to weights not included in the subnetwork. This gives us:
\begin{align*}
    &W_2(p(\rvw | \mathcal{D}), q_{S}(\rvw))^2\\
    &=W_2\left( \mathcal{N}(\widehat{\rvw}, \widetilde{\rmH}^{-1}),  \mathcal{N}(\widehat{\rvw}, \widetilde{\rmH}_{S+}^{-1} ) \right)^2\\
    &= \cancel{\| \widehat{\rvw} - \widehat{\rvw}\|_2^2} + \text{Tr}\left(\widetilde{\rmH}^{-1} + \widetilde{\rmH}_{S+}^{-1} - 2 \left(\widetilde{\rmH}_{S+}^{-1/2}\widetilde{\rmH}^{-1}\widetilde{\rmH}_{S+}^{-1/2}\right)^{1/2} \right)\\
    &= \text{Tr}\left(\widetilde{\rmH}^{-1} + \widetilde{\rmH}_{S+}^{-1} - 2 \left(\widetilde{\rmH}_{S+}^{-1/2}\widetilde{\rmH}^{-1}\widetilde{\rmH}_{S+}^{-1/2}\right)^{1/2} \right)\ .
\end{align*}

\subsection{Derivation for \cref{eq:wass2squared_indep}}
For $\widetilde{\rmH}^{-1} = \text{diag}(\sigma_1^2, \ldots, \sigma_D^2)$, the Wasserstein pruning objective in \cref{eq:wass2squared} simplifies to
\begin{align*}
    &W_2(p(\rvw | \mathcal{D}), q_{S}(\rvw))^2\\
    &=\text{Tr}\left(\widetilde{\rmH}^{-1}\right) + \text{Tr}\left(\widetilde{\rmH}_{S+}^{-1}\right) - 2\text{Tr}\left(\widetilde{\rmH}^{-1/2}\widetilde{\rmH}_{S+}^{-1/2}\right)\\
    &= \sum_{d=1}^D \sigma_d^2 + m_d \sigma_d^2 - 2 m_d \sigma_d^2 \\
    &= \sum_{d=1}^D \sigma_d^2(1 - m_d)\ ,
\end{align*}
where $m_d$ is the $d^{\text{th}}$ diagonal element of $\rmM_S$, i.e. $m_d = 1$ if $\rw_d$ is included in the subnetwork or 0 otherwise.

\FloatBarrier
\section{Updating the prior precision for uncertainty estimation with subnetworks}
\label{app:prior_update}

As described in \cref{sec:lin_laplace}, the linearised Laplace method can be understood as approximating our NN with a basis function linear model, where the jacobian of the NN evaluated at $\mathbf{x}$, $\mathbf{J}(\mathbf{x}) \in \mathcal{R}^{O \times D}$ represents the feature expansion. When employing an Isotropic Gaussian prior with precision $\lambda$ and for a given output dimension $i$, this formulation corresponds to a Gaussian process with kernel
\begin{gather} \label{eq:jacobian_kernel_expanded}
    k_{i}(\mathbf{x}, \mathbf{x}') =  \lambda^{-1} \mathbf{J}(\mathbf{x})_{i} \mathbf{J}(\mathbf{x}')_{i}^{\top} = \lambda^{-1} \sum_{d=1}^{D} \mathbf{J}(\mathbf{x})_{i,d} \mathbf{J}(\mathbf{x}')_{i,d}.
\end{gather}
For our subnetwork model, the Jacobian feature expansion is $\mathbf{J}_{S}(\mathbf{x}) \in \mathcal{R}^{O \times S}$, which is a submatrix of $\mathbf{J}(\mathbf{x})$. It follows that the implied kernel will be computed in the same way as \cref{eq:jacobian_kernel_expanded}, removing $D-S$ terms from the sum. 
The updated prior precision $\lambda_{S} = \lambda \! \cdot \! \nicefrac{S}{D}$ aims to maintain the magnitude of the sum, thus making the kernel corresponding to the subnetwork as similar as possible to that of the full network. 
\FloatBarrier
\section{Experimental Setup}
\label{sec:exp_setup}

\subsection{Toy Experiments}

We train a single, 2 hidden layer network, with 50 hidden ReLU units per layer using MAP inference until convergence. Specifically, we use SGD with a learning rate of $1 \times 10^{-3}$, momentum of $0.9$ and weight decay of $1 \times 10^{-4}$. We use a batch size of 512. The objective we optimise is the Gaussian log-likelihood of our data, where the mean is outputted by the network and the the variance is a hyperparameter learnt jointly with NN parameters by SGD. This variance parameters is shared among all datapoints. Once the network is trained, we perform post-hoc inference on it using different approaches. Since all of these involve the linearized approximation, the mean prediction is the same for all methods. Only their uncertainty estimates vary.

Note that while for this toy example, we could in principle use the full covariance matrix for the purpose of subnetwork selection, we still just use its diagonal (as described in \cref{sec:subnet_selection}) for consistency.
We use GGN Laplace inference over network weights (not biases) in combination with the linearized predictive distribution in \cref{eq:regression_predictive}.
Thus, all approaches considered share their predictive mean, allowing us to better compare their uncertainty estimates. 

All approaches share a single prior precision of $\lambda = 3$, scaled as $\lambda_{S} = \lambda \cdot \nicefrac{S}{D}$. We choose this prior precision such that the full covariance approach (optimistic baseline), where $\lambda_{S} = \lambda$, presents reasonable results. We first tried a precision of 1 and found the full covariance approach to produce excessively large errorbars (covering the whole plot). A value of 3 produces more reasonable results.

Final layer inference is performed by computing the full Laplace covariance matrix and discarding all entries except those corresponding to the final layer of the NN. 
Results for random sub-network selection are obtained with a single sample from a scaled uniform distribution over weight choice. 

\subsection{UCI Experiments}

In this experiment, our fully connected NNs have numbers of hidden layers $h_{d}{=}\{1, 2\}$ and hidden layer widths $w_{d}{=}\{50, 100\}$. For a dataset with input dimension $i_{d}$, the number of weights is given by $D{=}(i_{d}{+}1) w_{d}{+} (h_{d}{-}1)w_{d}^{2}$. Our 2 hidden layer, 100 hidden unit models have a weight count of the order $10^4$. The non-linearity used is ReLU.

We first obtain a MAP estimate of each model’s weights. Specifically, we use SGD with a learning rate of $1 \times 10^{-3}$, momentum of $0.9$ and weight decay of $1 \times 10^{-4}$. We use a batch size of 512. The objective we optimise is the Gaussian log-likelihood of our data, where the mean is outputted by the network and the the variance is a hyperparameter learnt jointly with NN parameters by SGD.

For each dataset split, we set aside 15\% of the train data as a validation set. We use these for early stopping training.
Training runs for a maximum of 2000 epochs but early stops with a patience of 500 if validation performance does not increase. For the larger Protein dataset, these values are 500 and 125. The weight settings which provide best validation performance are kept.

We then perform full network GGN Laplace inference for each model. We also use our proposed Wassertein rule together with the diagonal Hessian assumption to prune every network’s weight variances such that the number of variances that remain matches the size of every smaller network under consideration. The prior precision used for these steps is chosen such that the resulting predictor's logliklihood performance on the validation set is maximised. Specifically, we employ a grid search over the values: $\lambda: [0.0001, 0.001, 0.1, 0.5, 1, 2, 5, 10, 100, 1000]$.
In all cases, we employ the linearized predictive in \cref{eq:regression_predictive}. Consequently, networks with the same number of weights make the same mean predictions. Increasing the number of weight variances considered will thus only increase predictive uncertainty.

\subsection{Image Experiments}

The results shown in \cref{sec:img_exp} and \cref{sec:additional_results} are obtained by training ResNet-18 (and ResNet-50) models using SGD with momentum. For each experiment repetition, we train 7 different models: The first is for: `MAP', `Ours', `Ours (Rand)', `SWAG', `Diag-Laplace' and as the first element of `Ensemble'. We train 4 additional `Ensemble' elements, 1 network with `Dropout', and, finally 1 network for `VOGN'. The methods `Ours', `Ours (Rand)', `SWAG', and `Diag-Laplace' are applied post training.

For all methods except `VOGN' we use the following training procedure. The (initial) learning rate, momentum, and weight decay are 0.1, 0.9, and $1 \times 10^{-4}$, respectively. For `MAP' we use 4 Nvidia P100 GPUs with a total batch size of 2048. For the calculation of the Jacobian in the subnetwork selection phase we use a single P100 GPU with a batch size of 4. For the calculation of the hessian we use a single P100 GPU with a batch size of 2. We train on 1 Nvidia P100 GPU with a batch size of 256 for all other methods. Each dataset is trained for a different number of epochs, shown in \cref{tab:img_exp_train_configs}. We decay the learning rate by a factor of 10 at scheduled epochs, also shown in \cref{tab:img_exp_train_configs}. Otherwise, all methods and datasets share hyperparameters. These hyperparameter settings are the defaults provided by \texttt{PyTorch} for training on ImageNet. We found them to perform well across the board. 
We report results obtained at the final training epoch. We do not use a separate validation set to determine the best epoch as we found ResNet-18 and ResNet-50 to not overfit with the chosen schedules.

\begin{table}[htbp]
    \centering
    \caption{Per-dataset training configuration for image experiments.}
    \begin{tabular}{l|cc}
        \toprule
        \textsc{Dataset} & \textsc{No. Epochs} & \textsc{LR Schedule}  \\ \midrule
        MNIST & 90 & 40, 70 \\
        CIFAR10 & 300 & 150, 225 \\
        \bottomrule
    \end{tabular}
    \label{tab:img_exp_train_configs}
\end{table}

For `Dropout', we add dropout to the standard ResNet-50 model \citep{he2016deep} in between the \nth{2} and \nth{3} convolutions in the bottleneck blocks. This approach follows \citet{zagoruyko2016wide} and \citet{ashukha2020pitfalls} who add dropout in-between the two convolutions of a WideResNet-50's basic block. 
Following \citet{antoran2020depth}, we choose a dropout probability of 0.1, as they found it to perform better than the value of 0.3 suggested by \citet{ashukha2020pitfalls}. We use 16 MC samples for predictions.
`Ensemble' uses 5 elements for prediction. Ensemble elements differ from each other in their initialisation, which is sampled from the He initialisation distribution \citep{he2015delving}. We do not use adversarial training as, inline with \citet{ashukha2020pitfalls}, we do not find it to improve results.
For `VOGN' we use the same procedure and hyper-parameters as used by \citet{osawa2019} in their CIFAR10 experiments, with the exception that we use a learning rate of $1 \times 10^{-3}$ as we we found a value of $1 \times 10^{-4}$ not to result in convergence. We train on a single Nvidia P100 GPU with a batch size of 256. See the authors' GitHub for more details: \href{https://github.com/team-approx-bayes/dl-with-bayes/blob/master/distributed/classification/configs/cifar10/resnet18_vogn_bs256_8gpu.json}{\url{github.com/team-approx-bayes/dl-with-bayes/blob/master/distributed/classification/configs/cifar10/resnet18_vogn_bs256_8gpu.json}}.

We modify the standard ResNet-50 and ResNet-18 architectures such that the first $7\times7$ convolution is replaced with a $3\times3$ convolution. Additionally, we remove the first max-pooling layer.
Following \cite{goyal2017accurate}, we zero-initialise the last batch normalisation layer in residual blocks so that they act as identity functions at the start of training.

At test time, we tune the prior precision used for `Ours', `Diag-Laplace' and `SWAG' approximation on a validation set for each approach individually, as in \citet{ritter2018a, kristiadi2020being}. We use a grid search from $1 \times 10^{-4}$ to $1 \times 10^{4}$ in logarithmic steps, and then a second, finer-grained grid search between the two best performing values (again with logarithmic steps).

\subsection{Datasets}\label{app:data}

The 1d toy dataset used in \cref{sec:toy_1d} was taken from \cite{antoran2020depth}. We obtained it from the authors' github repo: \href{github.com/cambridge-mlg/DUN}{\url{https://github.com/cambridge-mlg/DUN}}. \cref{tab:tab_data} summarises the datasets used in \cref{sec:regression_UCI}.

We employ the Wine, Kin8nm and Protein datasets, together with their gap variants, because we find our models' performance to be most dependent on the quality of the estimated uncertainty here. On most other commonly used UCI regression datasets \citep{hernandez2015} we find increased uncertainty to hurt LL performance. In other words, the predictions made when using the MAP setting of the weights are better than those from any Bayesian ensemble.

Wine and Protein are available from the UCI dataset repository \cite{UCI_repo}. Kin8nm is available from \href{www.openml.org/d/189}{\url{https://www.openml.org/d/189}} \cite{foong2019between}.
For the standard splits~\citep{hernandez2015} 90\% of the data is used for training and 10\% for validation. For the gap splits~\citep{foong2019between} a split is obtained per input dimension by ordering points by their values across that dimension and removing the middle 33\% of the points. These are used for validation.

The datasets used for our image experiments are outlined in \cref{tab:image_datasets}.

\begin{table}[h]
\centering
\caption{Datasets from tabular regression used in \cref{sec:regression_UCI}}
\label{tab:tab_data}
\resizebox{\textwidth}{!}{\begin{tabular}{@{}ccccccccc@{}}
\toprule
Dataset     & N Train & N Val (15\% train) & N Test & Splits & Output Dim & Output Type & Input Dim & Input Type \\ \midrule
Wine        & 1223    & 216                & 160    & 20     & 1          & Continous   & 11        & Continous  \\
Wine Gap    & 906     & 161                & 532    & 11     & 1          & Continous   & 11        & Continous  \\
Kin8nm      & 6267    & 1106               & 819    & 20     & 1          & Continous   & 8         & Continous  \\
Kin8nm Gap  & 4642    & 820                & 2730   & 8      & 1          & Continous   & 8         & Continous  \\
Protein     & 34983   & 6174               & 4573   & 5      & 1          & Continous   & 9         & Continous  \\
Protein Gap & 25913   & 4573               & 15244  & 9      & 1          & Continous   & 9         & Continous  \\ \bottomrule
\end{tabular}}
\end{table}

\begin{table}[h]
    \centering
    \caption{Summary of image datasets. The test and train set sizes are shown in brackets, e.g. (test \& train).}
    \resizebox{\textwidth}{!}{\begin{tabular}{l|ccccc}
    \toprule
         \textsc{Name} & \textsc{Size} & \textsc{Input Dim.} & \textsc{No. Classes} & \textsc{No. Splits} \\
         \midrule
         MNIST~\citep{lecun1998gradient} & 70,000 (60,000 \& 10,000) & 784 ($28\,\times\,28$) & 10 & 2 \\
         Fashion-MNIST~\citep{xiao2017fashion} & 70,000 (60,000 \& 10,000) & 784 ($28\,\times\,28$) & 10 & 2 \\
         CIFAR10~\citep{krizhevsky2009learning} & 60,000 (50,000 \& 10,000) & 3072 ($32\,\times\,32\,\times\,3$) & 10 & 2 \\
         SVHN~\citep{netzer2011reading} & 99,289 (73,257 \& 26,032) & 3072 ($32\,\times\,32\,\times\,3$) & 10 & 2 \\
         \bottomrule
    \end{tabular}}
    \label{tab:image_datasets}
\end{table}

\end{document}